\pdfoutput=1
\documentclass[11pt]{article}

\usepackage[preprint]{acl}

\usepackage{times}
\usepackage{latexsym}

\usepackage[T1]{fontenc}

\usepackage[utf8]{inputenc}

\usepackage{microtype}

\usepackage{inconsolata}

\usepackage{graphicx}

\usepackage{amsmath}
\usepackage{amssymb}
\usepackage{multirow}
\usepackage{booktabs}
\usepackage[most]{tcolorbox}

\title{Speak vs. Know: Balancing Behavior and Knowledge\\Distributions of Large Language Models}

\author{
  \textbf{Yang Liu}\textsuperscript{\normalfont 1},
  \textbf{Masahiro Kaneko}\textsuperscript{\normalfont 2},
  \textbf{Chenhui Chu}\textsuperscript{\normalfont 1}
  \\
  \textsuperscript{1}Kyoto University,
  \textsuperscript{2}MBZUAI
  \\
  \texttt{yangliu@nlp.ist.i.kyoto-u.ac.jp} \\
  \texttt{Masahiro.Kaneko@mbzuai.ac.ae} \\
  \texttt{chu@i.kyoto-u.ac.jp}
}

\begin{document}
\maketitle
\begin{abstract}
  Recent work reports that the distribution verbalized directly in the output text aligns far better with human opinion distributions than the next-token log-probability distribution of LLMs.
  However, the verbalized distribution reflects what LLMs know about a group's opinions (\textbf{knowledge distribution}), while what they actually output at deployment is governed by their log-probability distribution (\textbf{behavior distribution}).
  We find that when prompting LLMs to predict the opinion of a specific group, the most likely answer under the verbalized distribution differs from that under the log-probability distribution for 37\% of World Values Survey questions, averaged over three LLMs. Thus, LLMs' knowledge does not match their behavior.
  This matters for evaluation: a verbalized distribution can score well against a group while its most likely answer is one the model never outputs, so the score no longer describes the model.
  The alignment scores for these two distributions reflect only the model's behavior or its knowledge, separately.
  We therefore propose a learnable combination of the two distributions that follows the most likely answer of the behavior distribution while preserving the shape of the knowledge distribution, with a single hyperparameter trading off behavioral fidelity against distribution quality.
  Our experiments on three opinion survey datasets and six models show that the balanced distribution improves behavioral consistency over the verbalized distribution from 71.6\% to 95.6\% on average, while the total variation distance to the human opinion distributions rises only slightly from 28.8\% to 33.8\%, and it remains effective under temperature scaling and cross-dataset settings.\footnote{Our code is available at \url{https://github.com/ku-nlp/speak-vs-know}.}
\end{abstract}

\section{Introduction}
\label{sec:intro}
Large language models (LLMs) are increasingly serving as an alternative to human opinion survey respondents:
to conduct a pre-test of a survey before launching a large-scale human survey~\citep{argyle2023out},
to simulate human behavior in agent-based social simulations~\citep{10.1145/3586183.3606763}, and to evaluate how well models represent the opinions of demographic and national groups~\citep{santurkar2023whose,liu2026alignment}.
In human opinion survey questions, no single answer is ``correct''; what matters is whether a model's output distribution matches a target group's opinion distribution, a property called \emph{distributional alignment}~\cite{meister-etal-2025-benchmarking}.

\begin{figure}[t]
  \centering
  \includegraphics[width=\columnwidth]{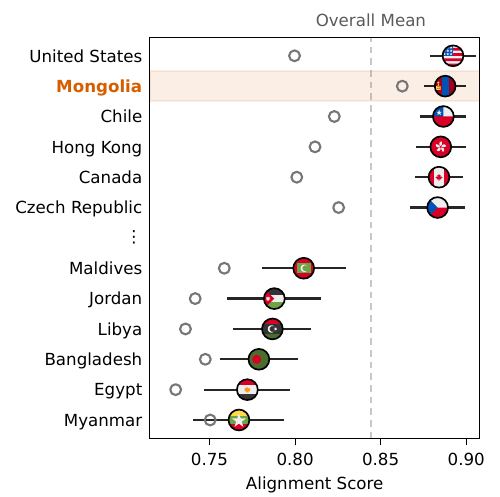}
  \caption{\textbf{The alignment score depends on the shape of the target.}
    Mongolia ranks second among the 66 countries and regions of the WVS in alignment with OLMo-3.1-32B-Instruct's verbalized distribution, under no steering, because its opinion distribution is nearly uniform;
  $\circ$ marks the alignment of each shown country and region's opinion distribution with the uniform distribution.}
  \label{fig:anomaly}
\end{figure}

\begin{figure*}[t]
  \centering
  \includegraphics[width=\textwidth]{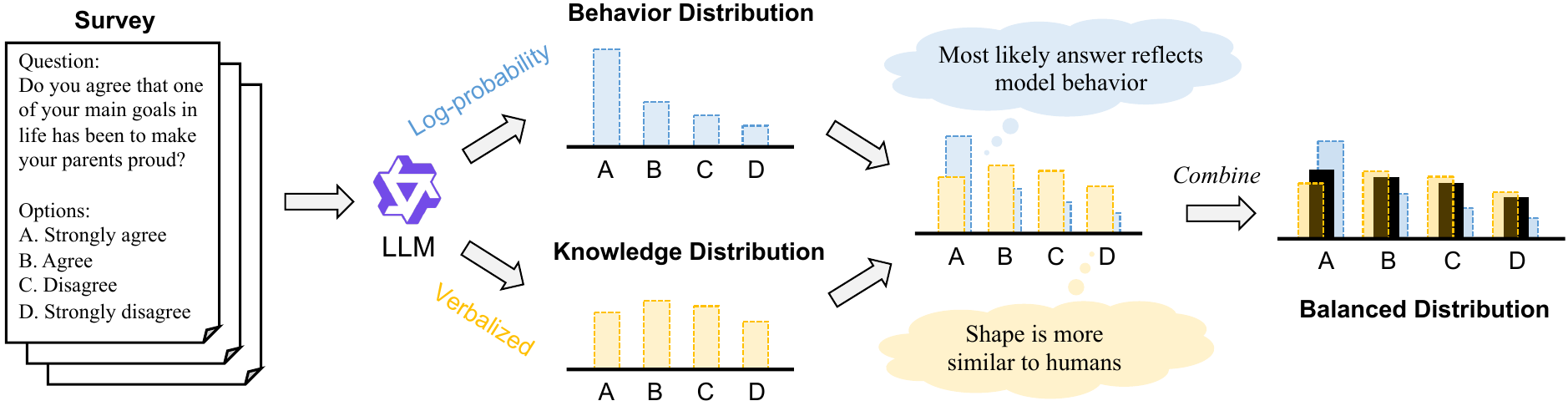}
  \caption{
  For each survey question, the LLM exposes a behavior distribution (log-probability) whose most likely answer reflects what the model actually outputs, and a knowledge distribution (verbalized) whose shape is closer to the human opinion distribution. We learn a combination of the two that keeps the most likely answer of the behavior distribution while preserving the shape of the knowledge distribution, yielding the balanced distribution.}
  \label{fig:method}
\end{figure*}

A model's opinion distribution can be expressed in two ways: \textbf{behavior distribution} and \textbf{knowledge distribution}.
The behavior distribution~\citep[e.g., log-probability distribution;][]{santurkar2023whose} refers to the next-token log-probabilities over option letters, which determine a model's output answer during inference due to greedy decoding.
The knowledge distribution~\citep[e.g., verbalized distribution;][]{meister-etal-2025-benchmarking} refers to the distribution that the model provides directly as a percentage list in textual format. This percentage list reflects the model's level of understanding of a group's opinions.
Regarding distributional alignment, the knowledge distribution is reported to be the clear winner~\cite{meister-etal-2025-benchmarking}: when simulating human opinion distribution, the shape of a model's verbalized distribution consistently aligns more closely with the shape of human opinion distribution than the log-probability distribution, which tends to be overly concentrated on one or two options \citep{meister-etal-2025-benchmarking,tian2023just}.

The verbalized distribution, however, also leads to some anomalous results.
As shown in Figure~\ref{fig:anomaly}, we rank the 66 countries and regions in the World Values Survey~\citep[WVS;][]{haerpfer2022wvs} by how closely OLMo-3.1-32B-Instruct's verbalized distribution aligns with their opinion distributions.
Mongolia is a low-resource language country, linguistically and culturally distant from the predominantly English, Western-centric data on which the model was trained.
Nevertheless, it ranks second, achieving alignment comparable to that of the United States and other Western countries.
In contrast, other linguistically distant countries, such as Myanmar and Egypt, occupy the lower end of the ranking.
Because OLMo-3.1-32B-Instruct has seen virtually no Mongolian text~\cite{soldaini2024dolma}, its alignment score with Mongolia is difficult to interpret.
We do not rule out the possibility that the model's opinions may actually align more closely with those of Mongolia, even though the model has barely seen any Mongolian corpora.

This uncertainty leads us to examine how the knowledge and behavior distributions differ.
A high alignment score only indicates that the shapes of the verbalized and human opinion distributions are similar.
It does not imply that the verbalized distribution reflects the most likely answer of the model.
Because the verbalized distribution is generated as text, there is no factor linking it to the probability of the next token that actually determines the model's answer.
In contrast, the log-probability distribution is constructed in a way that inherently predicts that answer. Under greedy decoding, the model outputs the most likely answer from its log-probability distribution.
Therefore, two distributions may disagree on the most likely answer.
When they do disagree, a metric based purely on distribution distance will favor a distribution that would never be observed when sampling from the model. Thus, neither expression method alone can serve as a faithful simulator.
In this study, we aim to construct a distribution that preserves the shape of the knowledge distribution while following the most likely answer of the behavior distribution.

Our study makes three major contributions.
\textit{1)}
We show that the two distributions capture different things: the verbalized distribution matches the shape of human opinion, while the log-probability distribution reflects what the model would answer (\S\ref{sec:motivation}).
\textit{2)} We introduce a learnable, behaviorally constrained mixing method to combine the knowledge and behavior distributions to obtain a balanced distribution (\S\ref{sec:method}).
\textit{3)} Finally, our experimental results show that the net utility obtained from the balanced distribution is higher than that from the knowledge distribution and the behavior distribution, and it outperforms temperature scaling settings (\S\ref{sec:results}).

\section{Motivation}
\label{sec:motivation}
We use the alignment score~\cite{santurkar2023whose, liu2026alignment} to investigate what the alignment score measures on the WVS dataset.
The alignment score is as follows:
\begin{equation}
\begin{split}
&\mathcal{A}(D_{\mathcal{M}},D_{c};\mathcal{Q}) \\
&\quad= \frac{1}{|\mathcal{Q}|}
\sum_{q \in \mathcal{Q}}
\left(
1 - \frac{\mathrm{WD} (D_{\mathcal{M}}(q), D_{c}(q) )}{|\mathcal{N}|-1}
\right)
\end{split}
\end{equation}
where $\mathcal{Q}$ is the question set, and $\mathcal{D}_{\mathcal{M}}(q)$, $\mathcal{D}_{c}(q)$ denote the opinion distributions of model $\mathcal{M}$ and country $c$ over the $|\mathcal{N}|$ ordered Likert-scale options of $q$. The factor $|\mathcal{N}|-1$ normalizes the Wasserstein distance ($\mathrm{WD}$)\footnote{\url{https://en.wikipedia.org/wiki/Wasserstein_metric}}, so that the alignment score lies in $[0,1]$, with 1 denoting a perfect match.
For each country, we compute the alignment score between a model's and the country's opinion distribution; the score ranges from 0 to 1, with higher values indicating better alignment.
Appendix~\ref{ssec:supp_prompts} provides the prompts, and Appendix~\ref{sec:supp_motivation} provides all numerical results of this section.

As shown in Table~\ref{tab:align_wvs},
the behavior distribution performs worse than the knowledge distribution under the persona steering setting.
In particular, the average alignment scores for the knowledge distribution range from 0.843 to 0.875, whereas the alignment scores for the behavior distribution range only from 0.765 to 0.804.
However, when the most likely answer of each distribution is compared with that of the human opinion distribution, the two are comparable in soft accuracy~\cite{alkhamissi2024investigating}.\footnote{Soft accuracy scores the most likely answer of a distribution against that of the human opinion
distribution as $1-|i_{\mathcal{M}}-i_{c}|/(|\mathcal{N}|-1)$, where $i_{\mathcal{M}}$ and $i_{c}$ are the indices of the two options on the Likert scale: an exact match scores 1, and the score decreases linearly with the distance between the options.}

\begin{table}[t]
\centering\small
\begin{tabular}{lrrrr}
\toprule
\multirow{2}{*}{\textbf{Model}} & \multicolumn{2}{c}{\textbf{Alignment score}} & \multicolumn{2}{c}{\textbf{Soft accuracy}} \\
\cmidrule(lr){2-3}\cmidrule(lr){4-5}
& Behav. & Knowl. & Behav. & Knowl. \\
\midrule
Qwen2.5-72B   & 0.765 & \textbf{0.875} & 0.799 & \textbf{0.809} \\
Llama-3.1-70B & 0.804 & \textbf{0.865} & \textbf{0.813} & 0.791 \\
OLMo-3.1-32B  & 0.779 & \textbf{0.843} & 0.748 & \textbf{0.765} \\
\bottomrule
\end{tabular}
\caption{The alignment score of the behavior (Behav.) and knowledge (Knowl.) distributions with the 66 WVS countries and regions under persona steering, and the soft accuracy of their most likely answers.
}
\label{tab:align_wvs}
\end{table}

Instead, the advantage lies in the shape: the dispersion of the knowledge distribution is far closer to that of the human opinion distribution, whereas the behavior distribution is over-concentrated.
As shown in Appendix~\ref{sec:supp_motivation} (Table~\ref{tab:supp_entropy_wvs}), the human opinion distribution on the WVS dataset is relatively flat (average entropy of 0.766), the knowledge distribution is even flatter (0.83 to 0.87), while the behavior distribution is much steeper (0.17 to 0.51).

The behavior distribution, however, determines the model's actual output: under greedy decoding, the model outputs the most likely answer from its log-probability distribution.
An ideal distribution should therefore reflect both the model's output behavior and its knowledge.

The knowledge distribution does not meet the first requirement, that of behavioral fidelity: its most likely answer often differs from that of the behavior distribution.
These two distributions share the same most likely answer for only 56.2\%–68.1\% of the questions (see Table~\ref{tab:bc_wvs}).

\begin{table}[t]
\centering\small
\begin{tabular}{lrr}
\toprule
\textbf{Model} & \textbf{No steering} & \textbf{Persona steering} \\
\midrule
Qwen2.5-72B   & 68.1\% & 64.9\% \\
Llama-3.1-70B & 61.1\% & 62.5\% \\
OLMo-3.1-32B  & 56.2\% & 60.5\% \\
\bottomrule
\end{tabular}
\caption{The behavioral consistency on the WVS: the percentage of questions on which the most likely answer of the knowledge distribution matches that of the behavior distribution.}
\label{tab:bc_wvs}
\end{table}

Therefore, when an LLM is steered to simulate a target group's opinions,
as shown in Figure~\ref{fig:method}, our goal is to create a single distribution that follows the most likely answer of the behavior distribution while preserving the shape of the knowledge distribution.

\section{Method}
\label{sec:method}

\subsection{Problem Setup}
\label{ssec:setup}
For a group $g$ and question $q$ with $n$ answer choices, let $\mathcal{D}_{gt}$ be the group's ground-truth answer distribution over the $n$ options.
We consider the following two distribution expression methods as behavior and knowledge distributions:
\begin{itemize}
\item Log-probability distribution $\mathcal{D}_{lp}$: the model's normalized next-token probabilities over the option letters (e.g., ``A'', ``B'', \dots);
\item Verbalized distribution $\mathcal{D}_{vb}$: the model is asked to verbalize the whole distribution in a JSON format of per-option percentages (e.g., \texttt{\{A:40\%, B:35\%, C:10\%, D:15\%\}});
\end{itemize}

As \S\ref{sec:motivation} shows, $\mathcal{D}_{lp}$ is behaviorally faithful but over-concentrated (low entropy), whereas $\mathcal{D}_{vb}$ captures the shape of a group's opinion but need not match the model's behavior.
Our goal is a single distribution $\mathcal{D}$ that balances the two types of distributions.

\subsection{Training Objective}
\label{ssec:objective}
We construct the balanced distribution $\mathcal{D}$ as a learnable combination of the behavior distribution and the knowledge distribution.
By adjusting its weights, we trade off between the shape of the distribution and the behavior fidelity. Then, we define the \textbf{balanced distribution} $\mathcal{D}$ and its two constraint terms: the \textbf{knowledge alignment term} $\mathcal{L}_\text{knowledge}$ and the \textbf{behavior constraint term} $\mathcal{L}_\text{behavior}$.
Finally, we integrate them into the training objective.

\paragraph{Balanced Distribution.}
By default, $\mathcal{D}$ uses a single weight to combine the log-probability and verbalized distributions:
\begin{equation}
\mathcal{D} = w\,\mathcal{D}_{lp} + (1-w)\,\mathcal{D}_{vb}, \qquad w\in[0,1],
\label{eq:bal}
\end{equation}
where $w{=}1$ recovers $\mathcal{D}_{lp}$ and $w{=}0$ recovers $\mathcal{D}_{vb}$, and the intermediate $w$ retains the shape of the verbalized distribution while pulling mass toward the model's actual output.
As both distributions are normalized and the weights are convex, $\mathcal{D}$ is automatically a valid distribution.

\paragraph{Knowledge Alignment Term.}
We quantify how much the balanced distribution differs in shape from the verbalized distribution $\mathcal{D}_{vb}$.
Specifically, we adopt the \emph{Total Variation} (TV) distance as our distributional distance measure, which compares the two distributions option by option:
\begin{equation}
\mathcal{L}_\text{knowledge}(\mathcal{D},\mathcal{D}_{vb}) = \frac{1}{2}\sum_{k=1}^{n}\big|\mathcal{D}^{(k)}-\mathcal{D}_{vb}^{(k)}\big|.
\label{eq:tv}
\end{equation}

The values for $\mathcal{L}_\text{knowledge}(\mathcal{D},\mathcal{D}_{vb})$ range from 0 to 1 and are lower when the balanced distribution $\mathcal{D}$ is closer to the verbalized distribution $\mathcal{D}_{vb}$.

\paragraph{Behavior Constraint Term.}
We aim to make the most likely answer of the balanced distribution as close as possible to the most likely answer under the behavior distribution.
Let $t=\arg\max_k \mathcal{D}_{lp}^{(k)}$ be the answer produced by the model at inference.
We require $\arg\max_k \mathcal{D}^{(k)} = t$ so that balancing never changes the model's most likely answer, and we enforce it with a soft hinge penalty:
\begin{equation}
\mathcal{L}_\text{behavior}(\mathcal{D}) = \max\!\big(0,\; \max_{k\neq t}\mathcal{D}^{(k)} - \mathcal{D}^{(t)} + m\big),
\label{eq:hinge}
\end{equation}
where $m\ge 0$ is a margin. When $t$ leads all other options by at least $m$, the penalty is zero; otherwise, the penalty increases linearly. Therefore, the larger $m$ is, the greater the lead required for the most likely answer.
We set $m{=}0$ throughout and vary only $\lambda$, so that the penalty is active exactly when $t$ is not the most likely answer of $\mathcal{D}$.
The effect of $m$ is analyzed in Appendix~\ref{sec:supp_margin}.

In addition, we quantify the consistency between a distribution and the behavior distribution. To this end, we introduce \emph{Behavioral Consistency} (BC), an evaluation metric that measures how often the mode of the knowledge distribution (or the balanced distribution) agrees with the mode of the behavior distribution.

\paragraph{Objective Function.}
We fit the weights of $\mathcal{D}$ by minimizing a single loss that combines the two terms:
\begin{equation}
\mathcal{L} = \underbrace{\mathcal{L}_\text{knowledge}(\mathcal{D}, \mathcal{D}_{vb})}_{\text{knowledge alignment}} \;+\; \lambda\, \underbrace{\mathcal{L}_\text{behavior}(\mathcal{D})}_{\text{behavior constraint}},
\label{eq:objective}
\end{equation}
where $\lambda$ is a constraint strength used to control the relative weights of $\mathcal{L}_\text{behavior}(\mathcal{D})$.
Eq.~\eqref{eq:objective} is the Lagrangian relaxation~\cite{geoffrion2009lagrangean} of minimizing $\mathcal{L}_\text{knowledge}$ subject to $\mathcal{L}_\text{behavior}\le\epsilon$.
Therefore, $\lambda$ is the Lagrange multiplier, the \emph{shadow price} of BC;
adjusting $\lambda$ controls the trade-off between the model's knowledge and behavior.

\subsection{Metrics}
\label{ssec:metrics}
In economics, a rational agent facing a decision weighs the benefit of an option
against its cost and chooses the option that maximizes the net utility, namely the benefit minus the cost~\cite{mas1995microeconomic}.
In this paper, we use net utility to describe the trade-off between knowledge and behavior:
\begin{equation}
\text{NU} = \text{BC} - \text{TV}.
\label{eq:utility}
\end{equation}
Here, TV is measured against the human opinion distribution, unlike $\mathcal{L}_\text{knowledge}$ in Eq.~\eqref{eq:tv}, which is measured against $\mathcal{D}_{vb}$. We report BC and TV as percentages and NU as a decimal.
For a balanced distribution, a higher BC indicates better alignment with the model's behavior, while a lower TV indicates better alignment with the human opinion distribution.
In this study, we consider BC and TV to be equally important, so their coefficients are both set to 1.
Then, the net utility $\text{NU}$ ranges from -1 to 1; a higher $\text{NU}$ indicates that the distribution better satisfies the requirements of both the model's behavior and the model's knowledge.

\section{Experimental Setup}
\label{sec:setup}

\paragraph{Datasets.}
We evaluate the three opinion survey datasets used in \citet{meister-etal-2025-benchmarking}'s work, which span demographic and national groups.
\textbf{OpinionQA}~\citep{santurkar2023whose} consists of 100 contentious PEW\footnote{\url{https://www.pewresearch.org/}} American Trends questions, spanning topics from science to politics and personal relationships, on which demographic subgroups frequently disagree; its ground truth is the answer distribution of six United States groups by party (\textit{Democrat} and \textit{Republican}), sex (\textit{Male} and \textit{Female}), and race (\textit{Black} and \textit{White}).
\textbf{NYT Book Opinions}~\citep{meister-etal-2025-benchmarking} captures non-political subjective preferences: 346 annotators give a 4-point Likert response to ``how likely are you to read it?'' for 235 books from the past two decades judged by \emph{The New York Times}, and the annotators are grouped by party (\textit{Democrat} and \textit{Republican}) and sex (\textit{Male} and \textit{Female}).
\textbf{GlobalOpinionQA}~\citep{durmus2023towards} is a global opinion dataset, taking its $100$ questions of highest cross-country disagreement from the WVS and the PEW Global Attitudes Survey, with each of the 26 countries forming a group. For every group and question pair, the ground truth is that group's normalized answer distribution from the original survey.

\paragraph{Models.}
We obtain the original verbalized and log-probability distributions using three popular open-source LLMs: OLMo-3.1-32B-Instruct\footnote{\url{https://huggingface.co/allenai/Olmo-3.1-32B-Instruct}}~\cite{olmo2025olmo}, Qwen2.5-72B-Instruct\footnote{\url{https://huggingface.co/Qwen/Qwen2.5-72B-Instruct}}~\cite{qwen2025qwen25technicalreport}, and Llama-3.1-70B-Instruct\footnote{\url{https://huggingface.co/meta-llama/Llama-3.1-70B-Instruct}}~\cite{grattafiori2024llama}.
These three LLMs come from three different developers.
In addition to open-source LLMs, we also validate our method on three production-grade APIs that support outputting the next-token log-probability: GPT-4o~\citep[\texttt{gpt-4o-2024-08-06};][]{openai2024gpt4ocard}, GPT-4.1~\citep[\texttt{gpt-4.1-2025-04-14};][]{gpt41}, and DeepSeek-V4-Flash~\citep[\texttt{deepseek-v4-flash};][]{xu2026deepseek}.

\paragraph{Steering Methods.}
We fit the balanced distribution only under steering, where the model is actually asked to simulate a target group. Without steering, the model returns the same distribution for every group, so there is no group-specific knowledge or behavior to balance.

We focus on two types of steering conditions that genuinely require LLMs to simulate target groups: 1) \textit{persona steering}, which instructs the model in the prompt to respond as a typical member of that group, and 2) \textit{few-shot steering}, which includes five additional context questions from the same group in the prompt, each with the group's actual opinion distribution to steer the model toward that group.
We create the few-shot demonstrations following \citet{meister-etal-2025-benchmarking}.
For the OpinionQA and NYT Book Opinions datasets, we keep the top five most similar questions based on the question similarity rankings in the benchmark.
Each question is accompanied by the corresponding ground truth distribution from that group.
For the GlobalOpinionQA dataset, as each country responds to only a small set of questions, we use five other questions answered by the same country as few-shot demonstrations.
For the few countries that answered fewer than five questions, we use all of the other answered questions.

\paragraph{Training Settings.}
We learn the weights for the balanced distribution by minimizing the objective function of Eq.~\eqref{eq:objective} using Adam~\cite{kingma2014adam} in full-batch training for 1,000 training steps, with a learning rate of $5\!\times\!10^{-2}$. In Appendix~\ref{sec:supp_convergence}, we provide loss and weight change curves to clarify that 1,000 steps are sufficient.
We assess generalization with grouped 5-fold cross-validation over questions:
for each dataset, we split it along the question dimension so that questions do not overlap between the training and validation sets.
This prevents data leakage.
For each fold, we fit $w$ based on the 4 training folds and report the average results of NU, TV, and BC across the 5 folds.

\section{Experimental Results}
\label{sec:results}

\subsection{Behavioral Fidelity at Near-Zero Cost}
\label{ssec:nearly_free}
\begin{figure}[t]
\centering
\includegraphics[width=0.7\columnwidth]{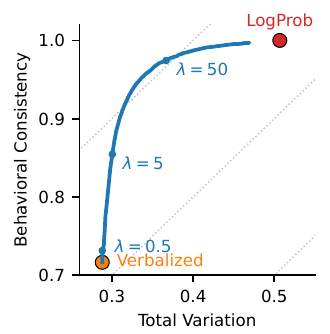}
\caption{Trade-offs between TV and BC under different $\lambda$. Average results across six models and three datasets, under persona steering.}
\label{fig:pareto}
\end{figure}

Fitting this balanced distribution requires balancing the knowledge alignment term
$\mathcal{L}_\text{knowledge}$ and the behavior constraint term $\mathcal{L}_\text{behavior}$ in Eq.~\eqref{eq:objective}.
As shown in Figure~\ref{fig:pareto}, sweeping $\lambda$ traces a curve in the TV and BC space.
At $\lambda = 0$ the behavior constraint is inactive and the balanced distribution recovers the verbalized distribution, so the left endpoint of the curve is the verbalized distribution itself, where BC is 71.6\%.\footnote{The 71.6\% averages the six models over the three evaluation datasets; Table~\ref{tab:bc_wvs} reports the results measured on the WVS.}
As $\lambda$ increases, the balanced distribution incorporates the log-probability distribution, raising BC while TV increases only slightly.

We select a single $\lambda^*$ for each model, shared across its three datasets: the $\lambda$ that maximizes the model's average net utility across the datasets.
Averaged over the six models and three datasets, BC reaches 95.6\% at this $\lambda^*$, while TV rises only slightly from 28.8\% at $\lambda=0$ to 33.8\%.
All results in this paper use this model-specific $\lambda^*$.
The weights $w$ are trained under $\lambda^*$ for each dataset.
Moreover, all $\lambda^*$ and weights $w$ are provided in Appendix~\ref{sec:supp_transfer} (Tables~\ref{tab:persona_weights} and~\ref{tab:fewshot_weights}).

Table~\ref{tab:main} compares the net utility $\text{NU}$ of the balanced distribution against each single distribution expression method under persona steering for Qwen2.5-72B-Instruct and GPT-4.1.
The balanced distribution achieves the highest $\text{NU}$ on all three datasets for both models, exceeding the strongest single method by 0.14 to 0.25.
On other models, the balanced distribution also outperforms all baseline distribution expression methods when averaged over the three datasets.

\begin{table}[t]
\centering
\small
\setlength{\tabcolsep}{6pt}
\begin{tabular}{lccc}
\toprule
\textbf{Method} & \textbf{OpinionQA} & \textbf{NYT Books} & \textbf{GOQA} \\
\midrule
\multicolumn{4}{l}{\emph{Qwen2.5-72B-Instruct (open-source)}} \\
LogProb   & 0.4791 & 0.3489 & 0.5002 \\
Verbalized & 0.4330 & 0.4794 & 0.4479 \\
Ours       & \textbf{0.6375} & \textbf{0.6196} & \textbf{0.6432} \\
\midrule
\multicolumn{4}{l}{\emph{GPT-4.1 (closed-source)}} \\
LogProb   & 0.4475 & 0.3008 & 0.5240 \\
Verbalized & 0.5460 & 0.3444 & 0.5297 \\
Ours       & \textbf{0.6946} & \textbf{0.5958} & \textbf{0.6926} \\
\bottomrule
\end{tabular}
\caption{The comparison of net utility $\text{NU}$ of each distribution when prompting the models using persona steering across three datasets.}
\label{tab:main}
\end{table}

In addition, we experimented by replacing the distribution distance metric TV with the Wasserstein distance~\cite{villani2009optimal} and substituting persona steering with few-shot steering. The conclusions are consistent with those based on TV and persona steering.
The full results are provided in Appendices~\ref{sec:supp_main_all} and~\ref{sec:supp_wass}.

\subsection{Robustness to Decoding Temperature}
\label{ssec:temp}
\citet{meister-etal-2025-benchmarking} demonstrated that direct temperature scaling of the log-probability distribution~\citep[TS-LogProb;][]{guo2017calibration}, i.e., $p_i(T) \propto p_i^{1/T}$, can change the sharpness of the model's log-probability distribution.
In particular, increasing the temperature can flatten the distribution.
Because temperature scaling is strictly monotonic, it does not change the model's most likely answer.
We select the deployable temperature parameters of LLMs $T \in \{0, 0.5, 1, 1.5, 2\}$.\footnote{The exact range allowed by the Generative API during deployment is $[0, 2]$, with some restrictions limited to $[0, 1]$.} We replace the log-probability distribution with its temperature scaled version.
Here, $T{=}0$ corresponds to the one-hot encoding distribution, while $T{=}1$ represents the identity mapping of the log-probability distribution.
We then retrain all weights from scratch under temperature scaling settings.

\begin{figure}[t]
\centering
\includegraphics[width=\columnwidth]{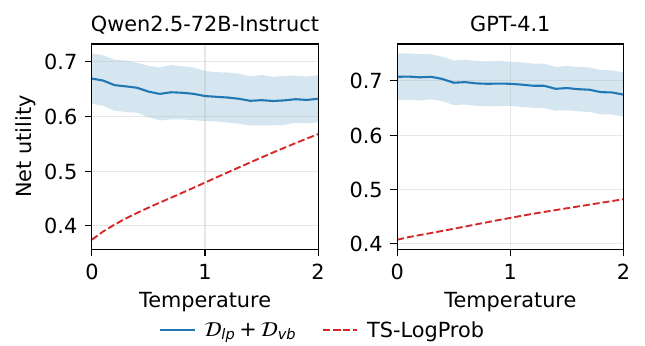}
\caption{The comparison of net utility between the balanced distribution and the temperature-scaled log-probability distribution at different temperatures.
The shadows indicate the standard deviation of the 5-fold cross-validation.}
\label{fig:temp_sweep}
\end{figure}

Figure~\ref{fig:temp_sweep} plots the net utility $\text{NU}$ of Qwen2.5-72B-Instruct and GPT-4.1 against the temperature $T$ on the OpinionQA dataset.
Two findings stand out in particular. First, the net utility of the learned balanced distribution remains nearly flat as temperature varies, suggesting that the balanced distribution requires behavioral signals from the log-probability distribution rather than its shape.
Second, the pure TS-LogProb is extremely sensitive to temperature: its net utility rises, averaged over the two panels, from 0.39 at $T{=}0.0$ to 0.53 at $T{=}2.0$, yet it never surpasses the balanced distribution at any temperature.
As shown above, our method is insensitive to temperature scaling and outperforms temperature scaling.

The results for the other two datasets and the remaining models, which show the same pattern, are provided in Appendix~\ref{sec:supp_temp}.

\subsection{Distribution Shape: Between Too-Flat and Too-Sharp}
\label{ssec:shape}
To characterize the shape of the balanced distribution, we compute the normalized entropy of each distribution, which ranges from $0$ (a sharp, one-hot shape) to $1$ (a uniform shape).
Table~\ref{tab:entropy} reports the absolute difference from the human entropy averaged over the three datasets under persona steering.
The log-probability distribution is severely over-concentrated (average entropy $0.32$), whereas the verbalized distribution is too flat ($0.85$), even flatter than the human opinion distribution ($0.75$).
The entropy of the balanced distribution almost perfectly matches that of the human opinion distribution ($0.73$ versus $0.75$), although the objective never optimizes entropy. For every one of the six models, the balanced distribution is the expression method closest to human entropy; it retains the shape information of the verbalized distribution while avoiding the log-probability distribution's tendency to concentrate on one or two options.
Thus, \textbf{the balanced distribution balances model behavior and knowledge}: its shape approximates the human opinion distribution while reflecting the model's actual behavioral choices.

\begin{table}[t]
\centering
\small
\setlength{\tabcolsep}{4pt}
\begin{tabular}{lrrr}
\toprule
\textbf{Model} & \textbf{LogProb} & \textbf{Verb.} & \textbf{Ours} \\
\midrule
Llama-3.1-70B & 0.286 & 0.080 & \textbf{0.045} \\
Qwen2.5-72B & 0.554 & 0.116 & \textbf{0.006} \\
OLMo-3.1-32B & 0.361 & 0.081 & \textbf{0.062} \\
GPT-4o & 0.458 & 0.113 & \textbf{0.022} \\
GPT-4.1 & 0.665 & 0.076 & \textbf{0.006} \\
DeepSeek-V4-F. & 0.253 & 0.103 & \textbf{0.008} \\
\midrule
Average & 0.429 & 0.095 & \textbf{0.025} \\
\bottomrule
\end{tabular}
\caption{Absolute difference between each distribution's normalized entropy and the human entropy ($0.751$), under persona steering, averaged over the three datasets (smaller means closer to the human opinion distribution). \textbf{Ours} is the closest to the human opinion distribution for every model. All open-source models are instruct models. F. indicates Flash.}
\label{tab:entropy}
\end{table}

\subsection{Cross-Dataset Validation}
\label{ssec:transfer}
To investigate whether the mixing weights $w$ in Eq.~\eqref{eq:bal}, which were trained on dataset $A$, also generalize to another dataset $B$, we conduct tests on the balanced distribution $\mathcal{D}_{lp} + \mathcal{D}_{vb}$.
Because a single $\lambda^*$ is shared across a model's datasets, only the weights $w$ are dataset specific.
For each source dataset $A$, we take the weights $w_A$ fitted in-domain on $A$ and apply them to each target dataset $B$; the resulting net utility is denoted as the transferred net utility.

\begin{figure}[t]
\centering
\includegraphics[width=\columnwidth]{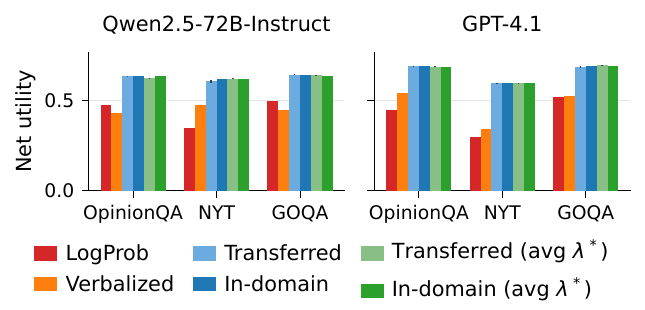}
\caption{The cross-dataset validation of the fitted weights $w$ at $\lambda^*$ for Qwen2.5-72B-Instruct and GPT-4.1. The weights fitted on one dataset still outperform all baseline distributions on another dataset. Blue bars use the $\lambda^*$ of each model; green bars use the average $\lambda^*$ of all six models.}
\label{fig:lambda_transfer}
\end{figure}

Figure~\ref{fig:lambda_transfer} shows the results for Qwen2.5-72B-Instruct and GPT-4.1.
For each target dataset, the transferred net utility converges to the in-domain net utility and exceeds that of the log-probability and verbalized distributions.
Among the results for all models in Appendix~\ref{sec:supp_transfer}, these transferred weights $w$ are effective. The only exception is DeepSeek-V4-Flash on the OpinionQA dataset, where the original log-probability distribution is already close to optimal ($\text{NU}=0.66$), leaving little space for improvement. Therefore, the effectiveness of the transferred weights $w$ enables generalization across different datasets.

Moreover, we replace each model's $\lambda^*$ with the average $\lambda^*$ of all six models.
As shown by the green bars in Figure~\ref{fig:lambda_transfer}, the net utility remains essentially unchanged, indicating that $\lambda^*$ itself need not be tuned for each model.

\section{Revisiting the Country Ranking}
\label{sec:mongolia}
In this section, we investigate two questions: how the balanced distribution changes the country ranking on the WVS, and whether the alignment scores are supported by an external measure of the model's cultural knowledge.
The weights for the balanced distribution are learned under persona steering and averaged over the three datasets.
These weights are then applied under no steering on the WVS.
The full results are provided in Appendix~\ref{sec:supp_mongolia}.

\paragraph{Country Ranking Changes.}
\begin{table}[t]
\centering
\setlength{\tabcolsep}{3.5pt}
\resizebox{\columnwidth}{!}{%
\begin{tabular}{lrr@{\hspace{1.0em}}lrr}
\toprule
\multicolumn{3}{c}{\textbf{Largest Falls}} & \multicolumn{3}{c}{\textbf{Largest Rises}} \\
\cmidrule(lr){1-3}\cmidrule(lr){4-6}
Country & Rank & $\Delta$ & Country & Rank & $\Delta$ \\
\midrule
Mongolia & 14.0 & $\downarrow$11.3 & Netherlands & 36.0 & $\uparrow$15.2 \\
Mexico & 27.3 & $\downarrow$9.0 & Germany & 28.2 & $\uparrow$12.2 \\
Thailand & 18.8 & $\downarrow$7.3 & New Zealand & 17.5 & $\uparrow$9.7 \\
Serbia & 35.2 & $\downarrow$6.7 & Philippines & 45.0 & $\uparrow$8.0 \\
Slovakia & 8.5 & $\downarrow$6.3 & Malaysia & 37.7 & $\uparrow$6.5 \\
\bottomrule
\end{tabular}}
\caption{The change in the country's ranking among the 66 countries when the balanced distribution is substituted for the verbalized distribution; this change is calculated as the average of the six models.
Rank is the country's average ranking under the verbalized distribution, and $\Delta$ is the number of places the country falls ($\downarrow$) or rises ($\uparrow$).}
\label{tab:ranking_changes}
\end{table}

Table~\ref{tab:ranking_changes} lists the countries with the largest changes in ranking when the balanced distribution is substituted for the verbalized distribution.
Among the 66 countries, Mongolia experiences the largest fall in ranking, dropping 11.3 places; under OLMo-3.1-32B-Instruct it falls from second place to tenth, dissolving the anomaly of \S\ref{sec:motivation}.
As shown in Appendix~\ref{sec:supp_mongolia} (Table~\ref{tab:supp_score_changes}), this adjustment is selective rather than uniform.
The alignment scores for English-speaking countries (e.g., the United States, Canada, and New Zealand) show minimal change (at most 0.012).
The largest declines occur for Mongolia and Slovakia, whose scores drop by 0.034 and 0.026 (up to 0.041 for Qwen2.5-72B-Instruct).

\paragraph{External Validation.}
\begin{table}[t]
\centering\small
\setlength{\tabcolsep}{3pt}
\resizebox{\columnwidth}{!}{%
\begin{tabular}{l ccc ccc}
\toprule
& \multicolumn{3}{c}{\textbf{MCQ}} & \multicolumn{3}{c}{\textbf{Short-Answer}} \\
\cmidrule(lr){2-4}\cmidrule(lr){5-7}
\textbf{Model} & LogProb & Verb. & Bal. & LogProb & Verb. & Bal. \\
\midrule
Llama-3.1-70B     & .622 & .621 & \textbf{.634} & .753 & .666 & \textbf{.759} \\
Qwen2.5-72B       & .648 & .596 & \textbf{.667} & .718 & .731 & \textbf{.791} \\
OLMo-3.1-32B      & .781 & .738 & \textbf{.787} & .863 & .796 & \textbf{.864} \\
GPT-4o            & \textbf{.525} & .508 & .494 & \textbf{.657} & .597 & .615 \\
GPT-4.1           & .500 & \textbf{.552} & .546 & .589 & .620 & \textbf{.621} \\
DeepSeek-V4-F. & \textbf{.550} & .507 & .516 & \textbf{.663} & .504 & .594 \\
\bottomrule
\end{tabular}}
\caption{The Pearson's $r$ between a country's BLEnD accuracy and its alignment score across the 11 overlapping countries, under the log-probability (LogProb), verbalized (Verb.), and balanced (Bal.) distributions.
\textbf{Bold} indicates the highest Pearson's $r$ of each model and task.}
\label{tab:blend}
\end{table}

If a balanced distribution can correct a certain artifact, its alignment score should reflect the model's actual understanding of each country.
We evaluate this on BLEnD~\citep{myung2024blend}, a benchmark dataset for general cultural knowledge that overlaps with 11 of the 66 WVS countries and regions.
For each model, we calculate the correlation between its BLEnD accuracy across countries on two tasks (multiple-choice questions (MCQ) and short-answer Q\&A) and the country's alignment score under each distribution expression method.
See Appendix~\ref{sec:supp_blend} for the complete experimental setup.
As shown in Table~\ref{tab:blend}, in 7 out of 12 cases, the balanced distribution outperforms the log-probability and verbalized distributions (10 of 12 against the verbalized distribution alone).
Therefore, a country's alignment scores under the balanced distribution better reflect the model's understanding of that country, which supports the ranking results of Table~\ref{tab:ranking_changes}.
It is worth noting that the log-probability distribution exhibits the highest Pearson's $r$ in 4 cases, which supports our claim that the log-probability distribution reflects the behavior of the model itself.

\section{Related Work}
\label{sec:related}

\paragraph{Simulating human opinions with LLMs.}
A growing number of studies are using LLMs to replace human survey respondents, both for piloting research and for populating agent-based simulations.
\citet{argyle2023out} introduced ``silicon samples'' as a way to use LLMs as cost-effective proxies for diverse demographic groups.
\citet{santurkar2023whose} curated OpinionQA from the Pew American
Trends Panel and showed that LLM responses systematically over-represent some demographics and under-represent others.
\citet{durmus2023towards} extended this to non-Western opinions with GlobalOpinionQA, exposing similar mis-representations.
Later studies report that LLM responses are unstable and diverge from human response biases \citep{liu2024robust, liu-2024-quantifying}, and that opinion probes are highly
sensitive to how questions are asked \citep{hida2025social, liu-chu-2025-llms, liu-chu-2026-understanding}. Steering LLMs toward a target group with personas or in-context examples can help narrow these gaps, but cannot eliminate them entirely \citep{hwang2023aligning, oba2024contextual, liu2026alignment}.

\paragraph{Eliciting probabilistic outputs from LLMs.}
There are two main methods for extracting distributions from LLMs: reading the next-token log-probabilities over the option tokens~\citep{santurkar2023whose} and verbalizing probabilities directly in the LLM's output~\citep{tian2023just, meister-etal-2025-benchmarking}.
\citet{tian2023just} showed that the verbalized distribution is better calibrated than log-probability distributions on correctness tasks, and \citet{kadavath2022language} found that models can assess the validity of their own answers. The closest phenomenon to ours is the generator-validator gap, where an LLM's generated answer contradicts its own judgment of that answer \citep{li2024benchmarking}. We study this tension in a new setting, subjective opinion distributions, and quantify the most likely answer disagreement between the verbalized and log-probability distributions elicited for the same group and question.

\section{Conclusion}
\label{sec:conclusion}
In this paper, we distinguished what an LLM knows about a group's opinions, captured by its verbalized distribution, from how it would actually behave at deployment, captured by its log-probability distribution.
Because the two disagreed on the most likely answer for a large fraction of questions, a high verbalized alignment score did not describe what a practitioner observed once the model was sampled, and a high alignment score did not, by itself, indicate that the distribution reflected the model.
Motivated by this, we proposed a learnable, behavior-constrained balanced distribution that preserves the shape of the knowledge distribution while anchoring its mode to the model's behavior, controlled by a single hyperparameter.
Across six models and three datasets, the balanced distribution consistently achieved higher net utility on average than the baseline distributions.
We recommend that future work on LLM opinion simulation report behavioral consistency alongside distributional alignment.

\section*{Acknowledgments}
This work was supported by JST BOOST, Grant Number JPMJBS2407.

\bibliography{custom}

\appendix

\section{More Settings}
\label{sec:supp_align_prompt}

\subsection{Alignment Score}
Following \citet{santurkar2023whose} and \citet{liu2026alignment}, the alignment score between a model distribution $\mathcal{D}$ and a ground-truth distribution $\mathcal{D}_{gt}$ over $n$ ordered options is as follows:
\begin{equation}
\mathcal{A}(\mathcal{D}, \mathcal{D}_{gt}) = 1 - \frac{1}{n-1}\sum_{k=1}^{n}\Big|\sum_{j\le k}\mathcal{D}^{(j)} - \sum_{j\le k}\mathcal{D}_{gt}^{(j)}\Big|,
\end{equation}
which ranges from 0 to 1, with higher values indicating better alignment.

\subsection{Prompts}
\label{ssec:supp_prompts}

\newtcolorbox{promptbox}[1]{enhanced, colback=gray!3, colframe=black!55, title={#1},
fonttitle=\bfseries\footnotesize, fontupper=\footnotesize\ttfamily, boxrule=0.5pt,
arc=1mm, left=1.5mm, right=1.5mm, top=1mm, bottom=1mm, before skip=8pt, after skip=8pt}

All prompts follow the structure of \citet{meister-etal-2025-benchmarking}: a task instruction, five formatting examples, and the question block (the question text followed by lettered options).
Under \textbf{persona steering}, the instruction asks the model to answer as the target group; the placeholder \texttt{<group>} below is instantiated as, e.g., ``a group of `Democrat''' for demographic groups or ``a typical respondent from Mongolia'' for countries.
Under \textbf{no steering}, the group phrase is simply omitted.
The instructions of the two distribution expression methods are shown below.

\begin{promptbox}{Log-probability instruction, persona steering}
I will provide a multiple choice response to a series of questions from a random group to help you understand the formatting of this task. After the 5 examples, please simulate an answer from \emph{<group>} for the question asked. Please only respond with a single multiple choice answer, no extra spaces, characters, quotes or text. Please only produce 1 character. Answers with more than one characters will not be accepted. First I will provide a few examples of this task just to give you a sense of the answer formatting, then you will perform the task yourself.
\end{promptbox}

\begin{promptbox}{Verbalized instruction, persona steering}
I will provide a distribution over answer choices on a series of questions from a random group to help you understand the formatting of this task. After the 5 examples, please express the distribution of answers from \emph{<group>} for the question asked. Do not follow the distribution expressed in the examples as they are only to give you an example of formatting, express the distribution based on your understanding of how humans may respond. Please only respond in the exact format of a dictionary mapping answer choice letter to probability, no extra spaces, characters, quotes or text. Please only produce 1 sentence in this format. Answers outside of this format will not be accepted.
\end{promptbox}

Under \textbf{few-shot steering}, the five formatting examples are replaced by up to five questions answered by the same group, each preceded by a context line that states the group's true response distribution. The prompt ends with a reminder before the target question.
The verbalized instruction is shown below; the log-probability variant differs only in the answer-format sentences, as above.

\begin{promptbox}{Verbalized instruction, few-shot steering}
In this task you will receive information on the distribution of responses from \emph{<group members>} to related questions. Given this data, your task is to simulate an answer to a new question from \emph{<group members>}. First, I will provide the distribution of responses from \emph{<group members>} to a series of questions. Afterwards, I will provide example responses to the question to help you understand the formatting of this task. After the examples, please express the distribution of answers from \emph{<group>} for the question asked. Please only respond in the exact format of a dictionary mapping answer choice letter to probability, no extra spaces, characters, quotes or text. Please only produce 1 sentence in this format. Answers outside of this format will not be accepted. For the new question, there will be no distribution provided, this is for you to estimate!\\
\\
Context: Let the probability that <group> responds to the following question with `<option 1>' be <percent>\%, `<option 2>' be <percent>\%, \ldots\ (one clause per answer option, covering the group's full response distribution).\\
\\
Your turn! Please answer this question for <group members>. As a reminder, this group is the exact same group as the group in the previous examples. The previous examples are used to provide an example of formatting and to give you insight into how this group would respond to such questions.
\end{promptbox}

Each question block lists the question text and its lettered options, and the model completes the final line after ``\texttt{Answer:}.''
The expected answer formats of the two methods are illustrated below with a WVS question.

\begin{promptbox}{Question block and answer formats (WVS example)}
Question: How important is family in your life?\\
A. Very important\\
B. Rather important\\
C. Not very important\\
D. Not at all important\\
\\
Log-probability:\ \ Answer: B\\
\textnormal{\footnotesize\itshape (the distribution is not taken from the generated text: it is read from the next-token probabilities of the option letters at the answer position, e.g., \{A: 0.21, B: 0.65, C: 0.10, D: 0.04\})}\\[4pt]
Verbalized:\ \ Answer: \{'A': '40\%', 'B': '35\%', 'C': '15\%', 'D': '10\%'\}\\
\textnormal{\footnotesize\itshape (the stated percentages are parsed directly as the distribution)}
\end{promptbox}

\subsection{Decoding Settings}
\label{ssec:supp_decoding}
The log-probability distribution is read from the first generated token: we request the top 20 token log-probabilities and limit the completion to one token.
The verbalized distribution is generated greedily ($T{=}0$); on the rare prompts where the output cannot be parsed as a distribution, it is re-sampled at increasing temperatures until a valid one is returned.
For the API models, the log-probability distribution is queried at $T{=}0$, except for DeepSeek-V4-Flash, which collapses to a one-hot distribution at $T{=}0$ and is therefore queried at $T{=}1$ to recover the raw next-token distribution.

\section{Details of the Motivation Analysis}
\label{sec:supp_motivation}
Tables~\ref{tab:supp_obs1a} and~\ref{tab:supp_obs1b} report the full per-model numbers behind the motivation analysis in Section~\ref{sec:motivation}: the average alignment scores of the two distribution expression methods under persona steering and the alignment scores with Mongolia, Myanmar, and the United States under no steering (including a uniform-distribution baseline).

\begin{table}[t]
\centering\small
\begin{tabular}{lrr}
\toprule
\textbf{Model} & \textbf{Behavior} & \textbf{Knowledge} \\
\midrule
Qwen2.5-72B-Instruct   & 0.765 & \textbf{0.875} \\
Llama-3.1-70B-Instruct & 0.804 & \textbf{0.866} \\
OLMo-3.1-32B-Instruct  & 0.779 & \textbf{0.843} \\
\bottomrule
\end{tabular}
\caption{Alignment score under persona steering, averaged over the 66 WVS countries. The knowledge distribution is the strongest single expression method, consistent with \citet{meister-etal-2025-benchmarking}.}
\label{tab:supp_obs1a}
\end{table}

\begin{table}[t]
\centering\small
\resizebox{\columnwidth}{!}{%
\begin{tabular}{lrrr}
\toprule
\textbf{Model} & \textbf{Mongolia} & \textbf{Myanmar} & \textbf{United States} \\
\midrule
Qwen2.5-72B-Instruct   & 0.879 & 0.783 & 0.905 \\
Llama-3.1-70B-Instruct & 0.875 & 0.772 & 0.906 \\
OLMo-3.1-32B-Instruct  & 0.888 & 0.767 & 0.892 \\
\textit{Uniform}  & \textit{0.863} & \textit{0.750} & \textit{0.800} \\
\bottomrule
\end{tabular}}
\caption{Alignment score with Mongolia, Myanmar, and the United States under no steering, plus a uniform-distribution baseline.}
\label{tab:supp_obs1b}
\end{table}

Section~\ref{sec:motivation} also compares the shapes of the two distributions.
Table~\ref{tab:supp_entropy_wvs} reports the normalized entropy behind that comparison: on the WVS, the human opinion distribution is relatively flat, the knowledge distribution is even flatter, and the behavior distribution is far more concentrated.
The normalized entropy ranges from 0 for a one-hot distribution to 1 for a uniform one.

\begin{table}[t]
\centering\small
\resizebox{\columnwidth}{!}{%
\begin{tabular}{lrr}
\toprule
\textbf{Model} & \textbf{Behavior} & \textbf{Knowledge} \\
\midrule
Qwen2.5-72B-Instruct   & 0.168 [0.161, 0.174] & 0.854 [0.848, 0.861] \\
Llama-3.1-70B-Instruct & 0.514 [0.503, 0.525] & 0.827 [0.822, 0.836] \\
OLMo-3.1-32B-Instruct  & 0.415 [0.397, 0.424] & 0.871 [0.866, 0.875] \\
\midrule
Human opinion & \multicolumn{2}{c}{0.766 [0.663, 0.885]} \\
\bottomrule
\end{tabular}}
\caption{The average normalized entropy of the behavior and knowledge distributions on the WVS dataset under no steering. The numbers in brackets are the minimum and maximum entropy over the 66 countries and regions.}
\label{tab:supp_entropy_wvs}
\end{table}

\section{Effect of the Margin}
\label{sec:supp_margin}
The behavior constraint term of Section~\ref{ssec:objective} includes a margin $m$, which sets how far the model's most likely answer $t$ must lead the remaining options before the penalty vanishes.
The main text fixes $m{=}0$ and varies only $\lambda$.
Here we justify that choice: $m$ and $\lambda$ move the balanced distribution along the same trade-off.

We scan $m \in \{0, 0.05, 0.1, 0.15, 0.2\}$ under two settings:
\begin{itemize}
\item $\lambda$ is held at the $\lambda^*$ fitted at $m{=}0$.
\item $\lambda$ is re-optimized for every value of $m$.
\end{itemize}
All numbers are averaged over the six models and the three datasets under persona steering.

\begin{figure}[t]
\centering
\includegraphics[width=\columnwidth]{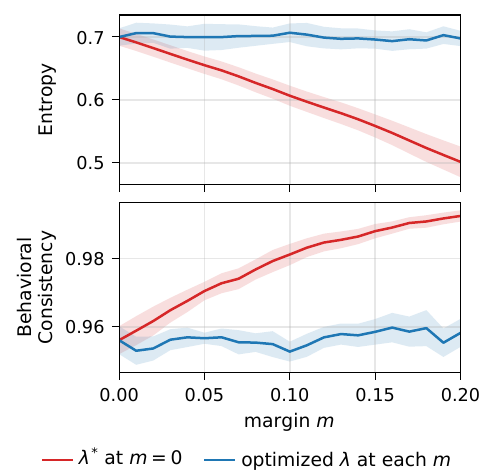}
\caption{Effect of the margin $m$ on the normalized entropy (top) and the behavioral consistency (bottom) of the balanced distribution, averaged over the six models and three datasets under persona steering. Red: $\lambda$ fixed at the value fitted at $m{=}0$; blue: $\lambda$ re-optimized at each $m$. The shaded areas are the standard error across the six models.}
\label{fig:supp_margin}
\end{figure}

Figure~\ref{fig:supp_margin} shows the result.
With $\lambda$ held fixed, raising $m$ from 0 to 0.2 sharpens the balanced distribution (its normalized entropy falls from $0.699$ to $0.501$) and raises behavioral consistency from $95.6\%$ to $99.2\%$.
A larger margin, therefore, gains behavioral fidelity at the cost of distribution shape, which is exactly the trade-off that $\lambda$ controls.
Once $\lambda$ is re-optimized at each $m$, both curves return to their $m{=}0$ values (at $m{=}0.2$, entropy $0.698$ and behavioral consistency $95.8\%$), and the fitted $\lambda^*$ absorbs the change instead, falling to $19\%$ of its value at $m{=}0$.

The margin thus adds no expressive power that $\lambda$ does not already provide: any shape reachable by increasing $m$ is also reachable by decreasing $\lambda$.
We keep $m{=}0$ so that a single hyperparameter governs the trade-off between distribution shape and behavioral consistency.

\section{Convergence of the Weight Fitting}
\label{sec:supp_convergence}
\begin{figure*}[t]
\centering
\includegraphics[width=0.8\textwidth]{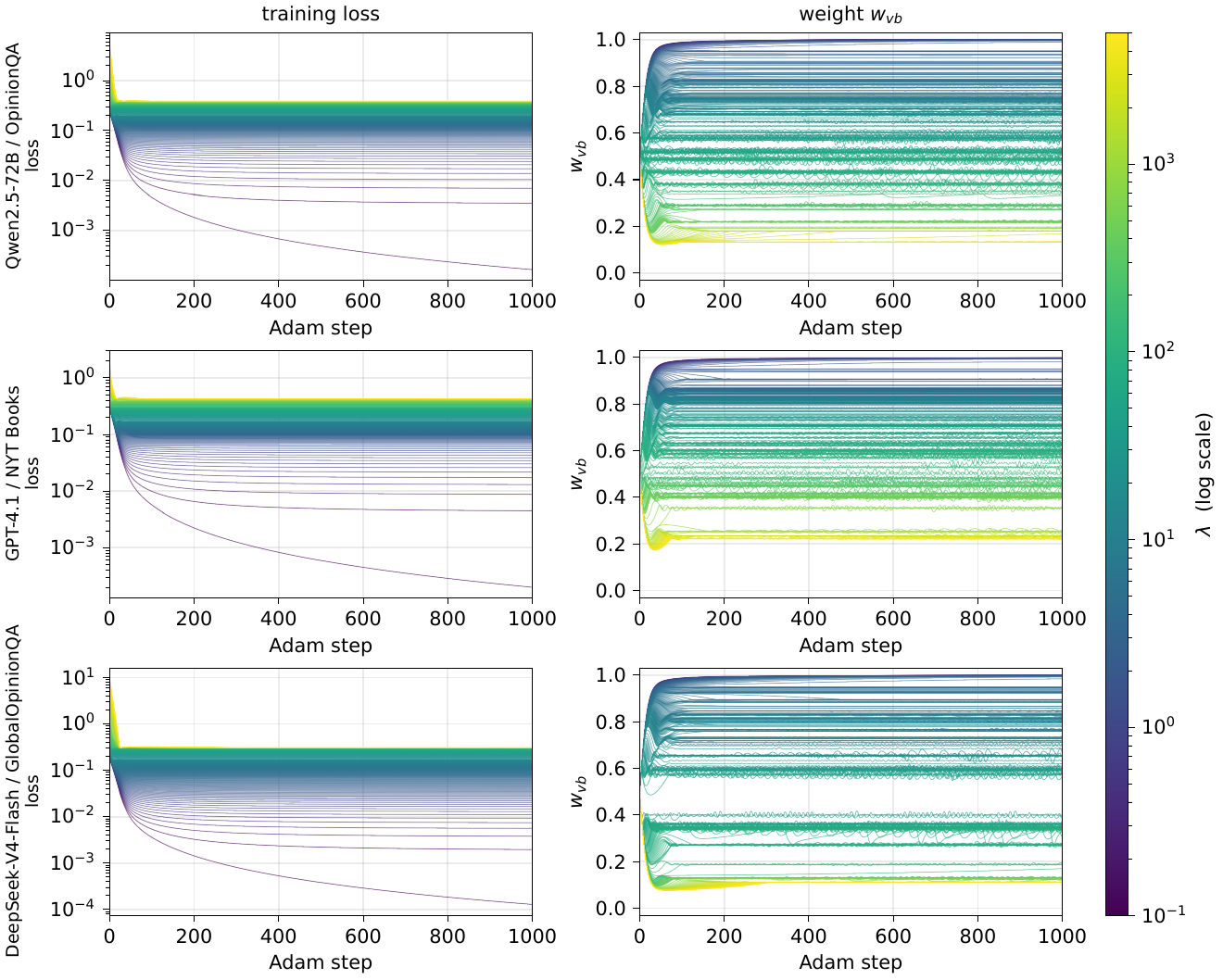}
\caption{Training loss (left) and the fitted verbalized weight $w_{vb}$ (right) at every one of the 1,000 Adam steps, for all values of $\lambda$ in the training grid (color: $\lambda$ on a log scale) and three model and dataset pairs under persona steering. Both the loss and the weight are flat long before the step budget ends.}
\label{fig:supp_convergence}
\end{figure*}

The weights of the balanced distribution are fitted with a fixed budget of 1,000 Adam steps and no early stopping, so the fit must converge within that budget for every $\lambda$ of the sweep.
Figure~\ref{fig:supp_convergence} instruments exactly that training loop and records the training loss of the objective function and the verbalized weight $w_{vb}$ at every step for all 321 values of $\lambda$ in the training grid and for three (model, dataset) pairs that cover the three datasets.

Both quantities flatten within the first few hundred steps.
Across the 321 values of $\lambda$, the loss enters and stays within $10^{-4}$ of its final value at a median step of 111 (Qwen2.5-72B / OpinionQA), 76 (GPT-4.1 / NYT Books), and 681 (DeepSeek-V4-Flash / GlobalOpinionQA).
The fitted weight is likewise frozen well before the budget ends: the change of $w_{vb}$ over the entire second half of training, $|w_{vb}^{(500)}-w_{vb}^{(1000)}|$, has a median of $8\!\times\!10^{-4}$ and a maximum of $2.2\!\times\!10^{-2}$ over all $\lambda$ and all three pairs, and the corresponding loss change is at most $1.9\!\times\!10^{-3}$.
The only trajectories that keep moving after step 500 are those at $\lambda\approx 0$ (the darkest curves), where the behavior constraint is inactive and the loss decays asymptotically toward the unconstrained optimum; these are also the trajectories whose weight has already saturated.
Training for 1,000 steps is therefore sufficient, and the reported results are insensitive to the exact step budget.

\section{Net Utility for All Models}
\label{sec:supp_main_all}
Tables~\ref{tab:supp_main_persona} and~\ref{tab:supp_main_fewshot} report the net utility of every distribution expression method for all six models under persona and few-shot steering, complementing the two models shown in Table~\ref{tab:main}.

\begin{table*}[t]
\centering\small
\begin{tabular}{ll rrr}
\toprule
\textbf{Model} & \textbf{Dataset} & \textbf{Log-prob} & \textbf{Verbalized} & \textbf{Ours} \\
\midrule
\multirow{3}{*}{Llama-3.1-70B} & OpinionQA & 0.6323 & 0.4420 & \textbf{0.6495} \\
& NYT Books & 0.4675 & 0.3254 & \textbf{0.5649} \\
& GOQA & 0.5980 & 0.4604 & \textbf{0.6222} \\
\midrule
\multirow{3}{*}{Qwen2.5-72B} & OpinionQA & 0.4791 & 0.4330 & \textbf{0.6375} \\
& NYT Books & 0.3489 & 0.4794 & \textbf{0.6196} \\
& GOQA & 0.5002 & 0.4479 & \textbf{0.6432} \\
\midrule
\multirow{3}{*}{OLMo-3.1-32B} & OpinionQA & 0.5033 & 0.3236 & \textbf{0.5387} \\
& NYT Books & 0.4078 & 0.3809 & \textbf{0.5356} \\
& GOQA & 0.4665 & 0.3171 & \textbf{0.5017} \\
\midrule
\multirow{3}{*}{GPT-4o} & OpinionQA & 0.5262 & 0.4578 & \textbf{0.6668} \\
& NYT Books & 0.4192 & 0.4392 & \textbf{0.5882} \\
& GOQA & 0.5470 & 0.5017 & \textbf{0.6652} \\
\midrule
\multirow{3}{*}{GPT-4.1} & OpinionQA & 0.4475 & 0.5460 & \textbf{0.6946} \\
& NYT Books & 0.3008 & 0.3444 & \textbf{0.5958} \\
& GOQA & 0.5240 & 0.5297 & \textbf{0.6926} \\
\midrule
\multirow{3}{*}{DeepSeek-V4-Flash} & OpinionQA & \textbf{0.6649} & 0.3650 & 0.6425 \\
& NYT Books & 0.4232 & 0.4241 & \textbf{0.5826} \\
& GOQA & 0.6219 & 0.4944 & \textbf{0.6837} \\
\bottomrule
\end{tabular}
\caption{Net utility $\mathrm{NU}=\mathrm{BC}-\mathrm{TV}$ of each distribution expression method under persona steering, for all six models.}
\label{tab:supp_main_persona}
\end{table*}

\begin{table*}[t]
\centering\small
\begin{tabular}{ll rrr}
\toprule
\textbf{Model} & \textbf{Dataset} & \textbf{Log-prob} & \textbf{Verbalized} & \textbf{Ours} \\
\midrule
\multirow{3}{*}{Llama-3.1-70B} & OpinionQA & 0.5631 & 0.3925 & \textbf{0.6304} \\
& NYT Books & 0.4855 & 0.3458 & \textbf{0.6870} \\
& GOQA & 0.6116 & 0.4105 & \textbf{0.6193} \\
\midrule
\multirow{3}{*}{Qwen2.5-72B} & OpinionQA & 0.4492 & 0.4326 & \textbf{0.6401} \\
& NYT Books & 0.3805 & 0.3201 & \textbf{0.7096} \\
& GOQA & 0.5316 & 0.4404 & \textbf{0.6349} \\
\midrule
\multirow{3}{*}{OLMo-3.1-32B} & OpinionQA & 0.4963 & 0.3449 & \textbf{0.5384} \\
& NYT Books & 0.4235 & 0.2033 & \textbf{0.6346} \\
& GOQA & 0.4973 & 0.2644 & \textbf{0.5357} \\
\midrule
\multirow{3}{*}{GPT-4o} & OpinionQA & 0.5073 & 0.4869 & \textbf{0.6757} \\
& NYT Books & 0.4992 & 0.3221 & \textbf{0.6608} \\
& GOQA & 0.5799 & 0.5383 & \textbf{0.6953} \\
\midrule
\multirow{3}{*}{GPT-4.1} & OpinionQA & 0.4305 & 0.5681 & \textbf{0.7072} \\
& NYT Books & 0.3517 & 0.3867 & \textbf{0.7077} \\
& GOQA & 0.5464 & 0.5993 & \textbf{0.7171} \\
\midrule
\multirow{3}{*}{DeepSeek-V4-Flash} & OpinionQA & 0.5711 & 0.4033 & \textbf{0.6432} \\
& NYT Books & 0.4980 & 0.3270 & \textbf{0.6357} \\
& GOQA & 0.6117 & 0.4600 & \textbf{0.6500} \\
\bottomrule
\end{tabular}
\caption{Net utility $\mathrm{NU}=\mathrm{BC}-\mathrm{TV}$ of each distribution expression method under few-shot steering, for all six models.}
\label{tab:supp_main_fewshot}
\end{table*}

\section{Net Utility under the Wasserstein Distance}
\label{sec:supp_wass}
The main text measures the distance between the model distribution and the ground-truth distribution with the total variation.
Here we replace it with the normalized Wasserstein distance $\mathrm{NWD}=1-\mathcal{A}$, where $\mathcal{A}$ is the alignment score of \S\ref{sec:supp_align_prompt}, i.e., the earth-mover distance over the ordinal answer options divided by $n-1$.
Unlike the total variation, this distance is ordinal: moving probability mass to a neighboring option is penalized less than moving it to a distant one.
The net utility becomes $\mathrm{NU}=\mathrm{BC}-\mathrm{NWD}$, the balanced distribution is re-fitted from scratch with the Wasserstein loss, and $\lambda^*$ is re-selected per model under the same rule as in Section~\ref{ssec:nearly_free}.
Because $\mathrm{NWD}\le\mathrm{TV}$ for ordinal options, all values are larger than in Tables~\ref{tab:supp_main_persona} and~\ref{tab:supp_main_fewshot}; only the comparison between the three columns is meaningful.

Tables~\ref{tab:supp_wass_persona} and~\ref{tab:supp_wass_fewshot} report the results.
The conclusion of Section~\ref{ssec:nearly_free} is unchanged.
The balanced distribution obtains the highest net utility on 13 of the 18 (model, dataset) combinations under persona steering and on 16 of 18 under few-shot steering; in every remaining case, the winner is the log-probability distribution and the gap is at most $0.007$, which is within the fold-to-fold spread of the cross-validation.
Averaged over the three datasets, the balanced distribution is the best expression method for all six models under both steering methods, by $0.006$ to $0.080$ (persona) and $0.019$ to $0.097$ (few-shot).
Compared with the total-variation results, the Wasserstein distance narrows the gap to the log-probability distribution, which is expected: the log-probability distribution concentrates its mass on a single option that is usually adjacent to the human mode, so an ordinal distance forgives its over-concentration far more than the total variation does.

\begin{table*}[t]
\centering\small
\begin{tabular}{ll c rrr}
\toprule
\textbf{Model} & \textbf{Dataset} & $\lambda^*$ & \textbf{Log-prob} & \textbf{Verbalized} & \textbf{Ours} \\
\midrule
\multirow{3}{*}{Llama-3.1-70B} & OpinionQA & \multirow{3}{*}{59} & \textbf{0.8327} & 0.5734 & 0.8300 \\
& NYT Books &  & 0.6690 & 0.4181 & \textbf{0.6941} \\
& GlobalOpinionQA &  & \textbf{0.8269} & 0.6156 & 0.8241 \\
\midrule
\multirow{3}{*}{Qwen2.5-72B} & OpinionQA & \multirow{3}{*}{18} & 0.7859 & 0.5772 & \textbf{0.8230} \\
& NYT Books &  & 0.6650 & 0.5688 & \textbf{0.7641} \\
& GlobalOpinionQA &  & 0.7991 & 0.6179 & \textbf{0.8354} \\
\midrule
\multirow{3}{*}{OLMo-3.1-32B} & OpinionQA & \multirow{3}{*}{44} & \textbf{0.8132} & 0.5010 & 0.8103 \\
& NYT Books &  & 0.6613 & 0.4912 & \textbf{0.6997} \\
& GlobalOpinionQA &  & \textbf{0.7774} & 0.5404 & 0.7764 \\
\midrule
\multirow{3}{*}{GPT-4o} & OpinionQA & \multirow{3}{*}{19} & 0.7769 & 0.5857 & \textbf{0.8189} \\
& NYT Books &  & 0.6833 & 0.5308 & \textbf{0.7389} \\
& GlobalOpinionQA &  & 0.7969 & 0.6446 & \textbf{0.8268} \\
\midrule
\multirow{3}{*}{GPT-4.1} & OpinionQA & \multirow{3}{*}{34} & 0.7603 & 0.6674 & \textbf{0.8487} \\
& NYT Books &  & 0.6325 & 0.4467 & \textbf{0.7375} \\
& GlobalOpinionQA &  & 0.8090 & 0.6647 & \textbf{0.8544} \\
\midrule
\multirow{3}{*}{DeepSeek-V4-Flash} & OpinionQA & \multirow{3}{*}{65} & \textbf{0.8331} & 0.5020 & 0.8262 \\
& NYT Books &  & 0.6524 & 0.5172 & \textbf{0.7082} \\
& GlobalOpinionQA &  & 0.8210 & 0.6339 & \textbf{0.8278} \\
\bottomrule
\end{tabular}
\caption{Net utility $\mathrm{NU}=\mathrm{BC}-\mathrm{NWD}$ of each distribution expression method under persona steering, for all six models, when the total variation is replaced by the normalized Wasserstein distance $\mathrm{NWD}=1-\mathcal{A}$. The balanced distribution is re-fitted with the Wasserstein loss and $\lambda^*$ is re-selected per model under the same rule.}
\label{tab:supp_wass_persona}
\end{table*}

\begin{table*}[t]
\centering\small
\begin{tabular}{ll c rrr}
\toprule
\textbf{Model} & \textbf{Dataset} & $\lambda^*$ & \textbf{Log-prob} & \textbf{Verbalized} & \textbf{Ours} \\
\midrule
\multirow{3}{*}{Llama-3.1-70B} & OpinionQA & \multirow{3}{*}{24} & 0.8158 & 0.5247 & \textbf{0.8207} \\
& NYT Books &  & 0.7542 & 0.4370 & \textbf{0.8181} \\
& GlobalOpinionQA &  & \textbf{0.8385} & 0.5719 & 0.8323 \\
\midrule
\multirow{3}{*}{Qwen2.5-72B} & OpinionQA & \multirow{3}{*}{47} & 0.7800 & 0.5761 & \textbf{0.8321} \\
& NYT Books &  & 0.7174 & 0.4091 & \textbf{0.8280} \\
& GlobalOpinionQA &  & 0.8104 & 0.6062 & \textbf{0.8365} \\
\midrule
\multirow{3}{*}{OLMo-3.1-32B} & OpinionQA & \multirow{3}{*}{42} & \textbf{0.8010} & 0.5093 & 0.7989 \\
& NYT Books &  & 0.7091 & 0.3031 & \textbf{0.7824} \\
& GlobalOpinionQA &  & 0.7884 & 0.4680 & \textbf{0.7954} \\
\midrule
\multirow{3}{*}{GPT-4o} & OpinionQA & \multirow{3}{*}{28} & 0.7787 & 0.6194 & \textbf{0.8366} \\
& NYT Books &  & 0.7487 & 0.4134 & \textbf{0.7968} \\
& GlobalOpinionQA &  & 0.8169 & 0.6750 & \textbf{0.8506} \\
\midrule
\multirow{3}{*}{GPT-4.1} & OpinionQA & \multirow{3}{*}{44} & 0.7571 & 0.6928 & \textbf{0.8499} \\
& NYT Books &  & 0.6945 & 0.4778 & \textbf{0.8355} \\
& GlobalOpinionQA &  & 0.8150 & 0.7204 & \textbf{0.8711} \\
\midrule
\multirow{3}{*}{DeepSeek-V4-Flash} & OpinionQA & \multirow{3}{*}{31} & 0.8125 & 0.5402 & \textbf{0.8168} \\
& NYT Books &  & 0.7367 & 0.4317 & \textbf{0.7865} \\
& GlobalOpinionQA &  & 0.8275 & 0.5990 & \textbf{0.8318} \\
\bottomrule
\end{tabular}
\caption{Net utility $\mathrm{NU}=\mathrm{BC}-\mathrm{NWD}$ of each distribution expression method under few-shot steering, for all six models, when the total variation is replaced by the normalized Wasserstein distance $\mathrm{NWD}=1-\mathcal{A}$. The balanced distribution is re-fitted with the Wasserstein loss and $\lambda^*$ is re-selected per model under the same rule.}
\label{tab:supp_wass_fewshot}
\end{table*}

\section{Temperature Scaling for All Models and Datasets}
\label{sec:supp_temp}
\begin{figure*}[t]
\centering
\includegraphics[width=0.8\textwidth]{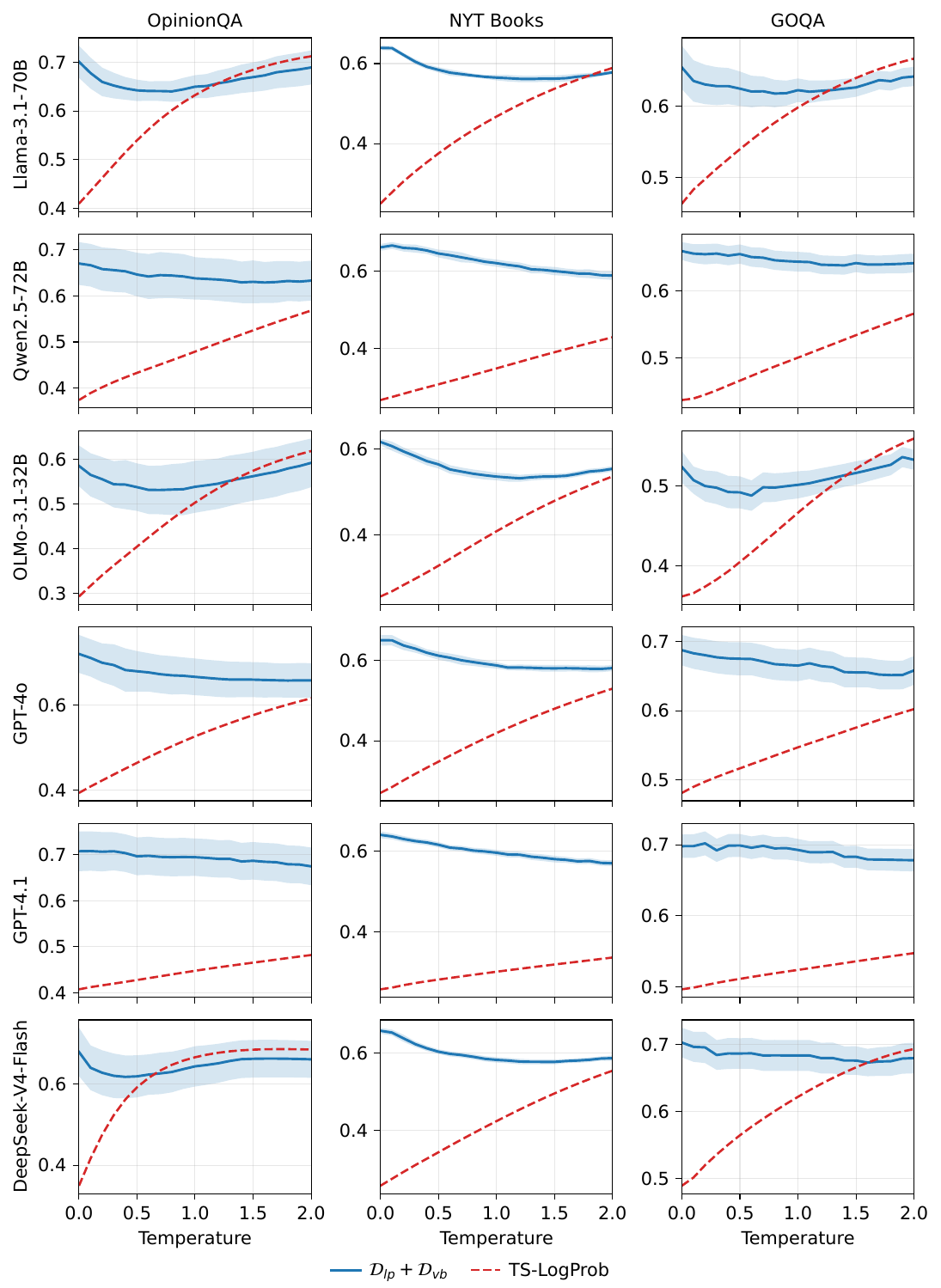}
\caption{The comparison of net utility between the balanced
distribution and the temperature-scaled log-probability distribution at different temperatures. The shadows indicate the standard deviation of the 5-fold cross-validation.}
\label{fig:supp_temp}
\end{figure*}

Figure~\ref{fig:supp_temp} extends the temperature analysis of Section~\ref{ssec:temp} to all six models and all three datasets. The overall pattern holds: the balanced distribution is nearly flat in temperature, whereas the temperature-scaled log-probability distribution is highly sensitive. On a few open-weight pairs (e.g., Llama-3.1-70B-Instruct and DeepSeek-V4-Flash on OpinionQA), the temperature-scaled distribution catches up with the balanced distribution at the highest temperatures, but it does not surpass it by a meaningful margin, and on NYT Books, it remains clearly below throughout.

\section{Cross-Dataset Transfer for All Models}
\label{sec:supp_transfer}
\begin{figure*}[t]
\centering
\includegraphics[width=\textwidth]{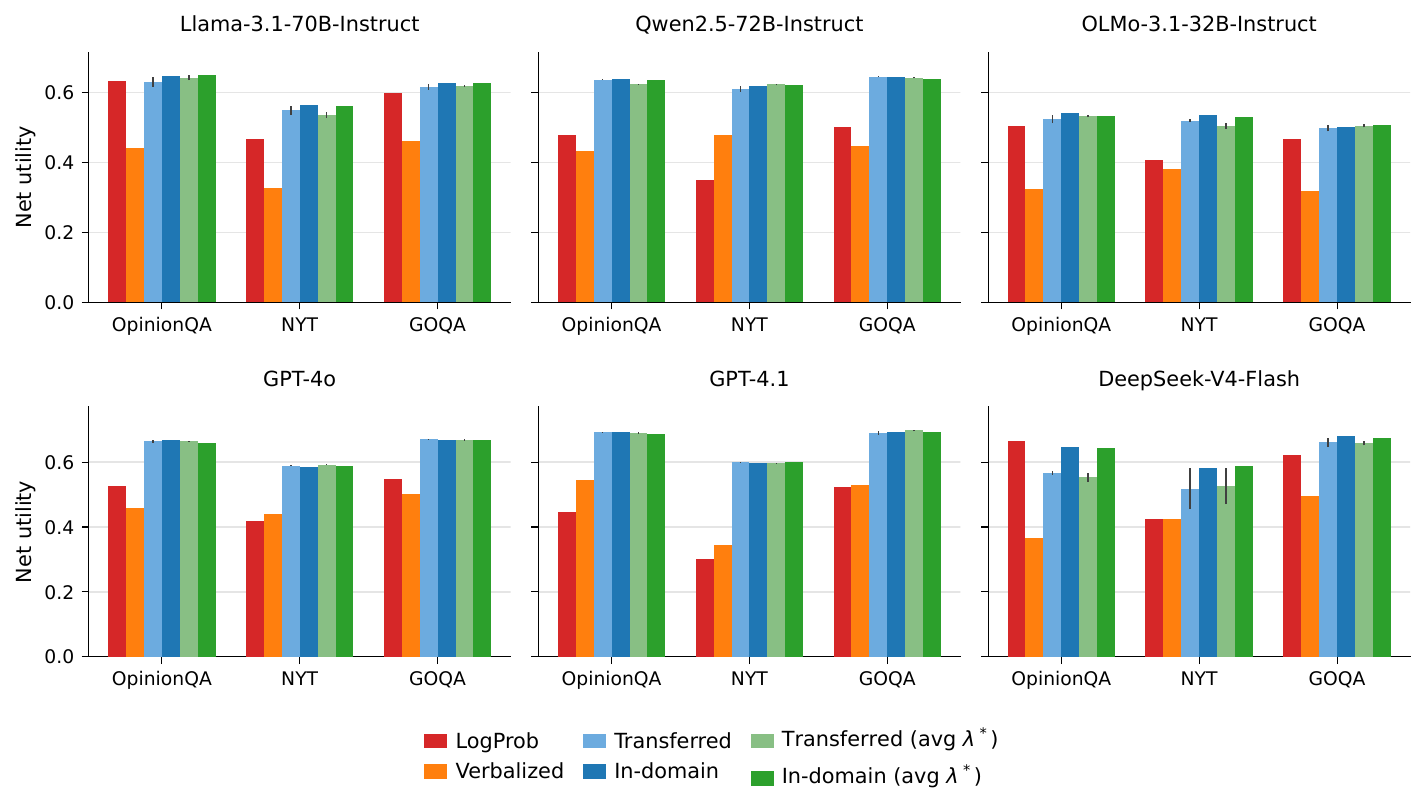}
\caption{Cross-dataset transfer of the fitted weights $w$ (at the per-model $\lambda^{*}$) for all six models under persona steering. Blue bars use the $\lambda^*$ of each model; green bars use the average $\lambda^*$ of all six models.}
\label{fig:supp_transfer_persona}
\end{figure*}

\begin{figure*}[t]
\centering
\includegraphics[width=\textwidth]{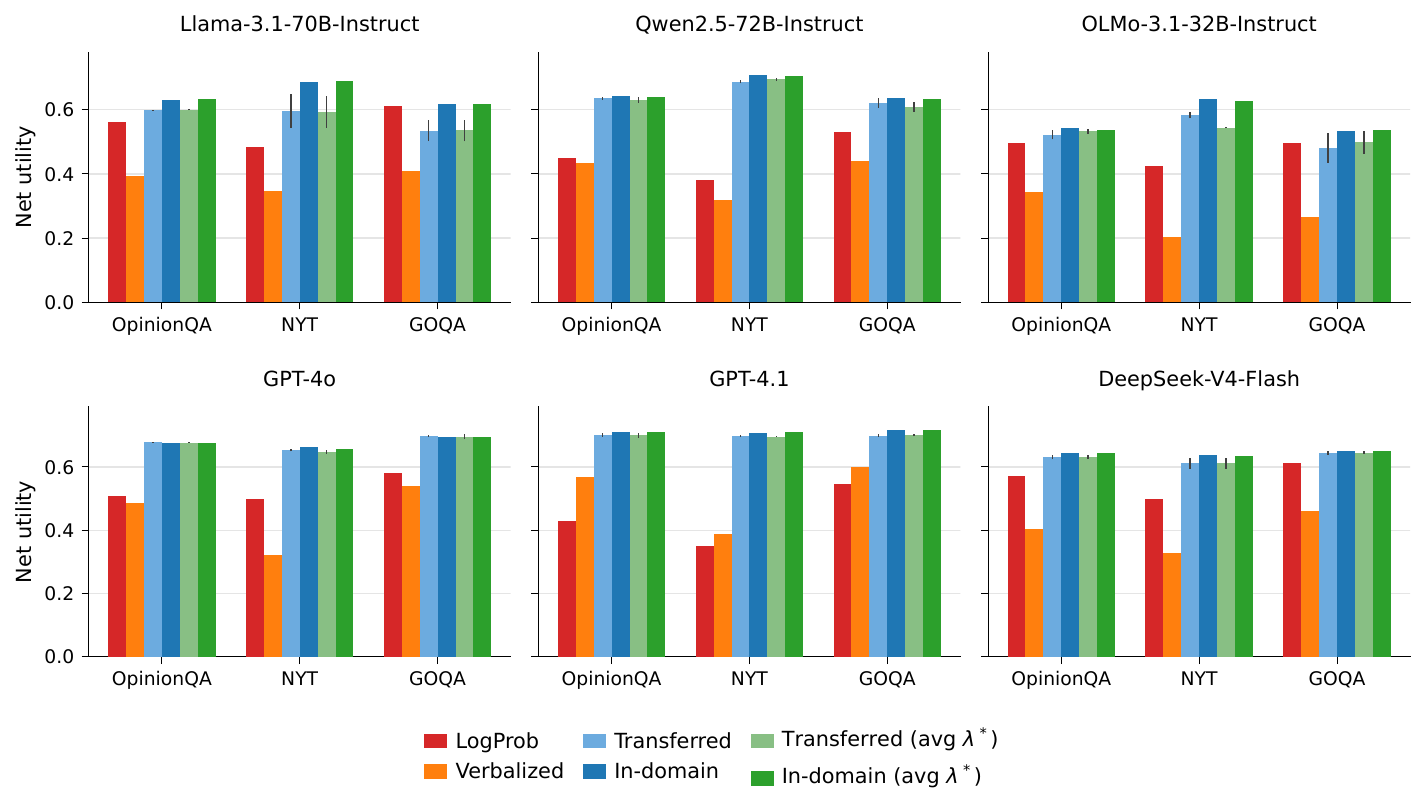}
\caption{Cross-dataset transfer of the fitted weights $w$ (at the per-model $\lambda^{*}$) for all six models under few-shot steering. Blue bars use the $\lambda^*$ of each model; green bars use the average $\lambda^*$ of all six models.}
\label{fig:supp_transfer_fewshot}
\end{figure*}

Tables~\ref{tab:persona_weights} and~\ref{tab:fewshot_weights} report the learned $\lambda^*$ and weights.
Figures~\ref{fig:supp_transfer_persona} and~\ref{fig:supp_transfer_fewshot} extend the cross-dataset validation of Section~\ref{ssec:transfer} to all six models under persona and few-shot steering. The transferred net utility converges to the in-domain net utility and surpasses all baseline distributions on nearly every target; the only exception is DeepSeek-V4-Flash on OpinionQA, where the log-probability distribution is already close to optimal. Moreover, replacing the per-model $\lambda^*$ with the average $\lambda^*$ of all six models (green bars) leaves the net utility essentially unchanged, indicating that $\lambda^*$ itself need not be tuned for each model.

\begin{table*}[t]
\centering\small
\begin{tabular}{l c cc cc cc}
\toprule
\multirow{2}{*}{\textbf{Model}} & \multirow{2}{*}{$\lambda^*$} & \multicolumn{2}{c}{\textbf{OpinionQA}} & \multicolumn{2}{c}{\textbf{NYT Books}} & \multicolumn{2}{c}{\textbf{GlobalOpinionQA}} \\
\cmidrule(lr){3-4}\cmidrule(lr){5-6}\cmidrule(lr){7-8}
& & $w_{lp}$ & $w_{vb}$ & $w_{lp}$ & $w_{vb}$ & $w_{lp}$ & $w_{vb}$ \\
\midrule
Llama-3.1-70B & 18 & 0.70 & 0.30 & 0.43 & 0.57 & 0.58 & 0.42 \\
Qwen2.5-72B & 37 & 0.41 & 0.59 & 0.35 & 0.65 & 0.36 & 0.64 \\
OLMo-3.1-32B & 16 & 0.61 & 0.39 & 0.45 & 0.55 & 0.58 & 0.42 \\
GPT-4o & 38 & 0.40 & 0.60 & 0.44 & 0.56 & 0.38 & 0.62 \\
GPT-4.1 & 16 & 0.24 & 0.76 & 0.20 & 0.80 & 0.23 & 0.77 \\
DeepSeek-V4 & 34 & 0.89 & 0.11 & 0.37 & 0.63 & 0.39 & 0.61 \\
\bottomrule
\end{tabular}
\caption{
The per-model $\lambda^*$ (a single $\lambda$ per model, shared across its three datasets, that maximizes the model\textquotesingle s average net utility) and the corresponding mixing weights $w_{lp}, w_{vb}$ of $\mathcal{D}_{lp}+\mathcal{D}_{vb}$ under persona steering.}
\label{tab:persona_weights}
\end{table*}

\begin{table*}[t]
\centering\small
\begin{tabular}{l c cc cc cc}
\toprule
\multirow{2}{*}{\textbf{Model}} & \multirow{2}{*}{$\lambda^*$} & \multicolumn{2}{c}{\textbf{OpinionQA}} & \multicolumn{2}{c}{\textbf{NYT Books}} & \multicolumn{2}{c}{\textbf{GlobalOpinionQA}} \\
\cmidrule(lr){3-4}\cmidrule(lr){5-6}\cmidrule(lr){7-8}
& & $w_{lp}$ & $w_{vb}$ & $w_{lp}$ & $w_{vb}$ & $w_{lp}$ & $w_{vb}$ \\
\midrule
Llama-3.1-70B & 20 & 0.54 & 0.46 & 0.31 & 0.69 & 0.84 & 0.16 \\
Qwen2.5-72B & 30 & 0.31 & 0.69 & 0.22 & 0.78 & 0.36 & 0.64 \\
OLMo-3.1-32B & 12.1 & 0.59 & 0.41 & 0.33 & 0.67 & 0.65 & 0.35 \\
GPT-4o & 25 & 0.29 & 0.71 & 0.37 & 0.63 & 0.32 & 0.68 \\
GPT-4.1 & 20 & 0.18 & 0.82 & 0.21 & 0.79 & 0.30 & 0.70 \\
DeepSeek-V4 & 20 & 0.53 & 0.47 & 0.44 & 0.56 & 0.70 & 0.30 \\
\bottomrule
\end{tabular}
\caption{
The per-model $\lambda^*$ (a single $\lambda$ per model, shared across its three datasets, that maximizes the model\textquotesingle s average net utility) and the corresponding mixing weights $w_{lp}, w_{vb}$ of $\mathcal{D}_{lp}+\mathcal{D}_{vb}$ under few-shot steering.}
\label{tab:fewshot_weights}
\end{table*}

\section{Country Ranking for All Models}
\label{sec:supp_mongolia}
Section~\ref{sec:mongolia} re-scores the 66 WVS countries and regions with the balanced distribution and reports the largest ranking changes.
Here we report the complete picture: for every one of the six models, all 66 WVS countries and regions are re-scored, and we compare how they are \emph{ranked} by the verbalized distribution and by the balanced distribution.
The setting is that of Section~\ref{sec:mongolia}: the $\mathcal{D}_{lp}+\mathcal{D}_{vb}$ mixing weights are learned under persona steering at each model's $\lambda^*$ (Table~\ref{tab:persona_weights}), averaged over the three datasets, and then applied unchanged to the WVS no-steering distributions.
Table~\ref{tab:supp_mongolia_rank} lists every country's alignment score and its rank under both distributions.

Table~\ref{tab:supp_score_changes} summarizes the alignment score changes mentioned in Section~\ref{sec:mongolia} for seven representative countries averaged over the six models.
The adjustment is selective rather than uniform.
English-speaking countries barely move (at most $0.012$, and New Zealand is unchanged), and Hong Kong, as a partially English-speaking region, sees a slight decrease ($0.016$).
The largest declines occur for Mongolia and Slovakia ($0.034$ and $0.026$; up to $0.041$ for Qwen2.5-72B-Instruct), while Egypt, whose verbalized score is already the lowest, drops by only about half as much as Mongolia.
The size of the decrease therefore measures how much of a country's alignment score rests on a shape that the model's own behavior does not support.

\begin{table}[t]
\centering\small
\setlength{\tabcolsep}{8pt}
\begin{tabular}{l cc}
\toprule
\multirow{2}{*}{\textbf{Country}} & \multicolumn{2}{c}{\textbf{Alignment Score}} \\
\cmidrule(lr){2-3}
& Verbalized & Balanced \\
\midrule
United States & 0.910 & 0.898\,\textsubscript{$\downarrow$.012} \\
Canada        & 0.894 & 0.890\,\textsubscript{$\downarrow$.004} \\
New Zealand   & 0.879 & 0.879\,\textsubscript{$\downarrow$.000} \\
Hong Kong     & 0.883 & 0.867\,\textsubscript{$\downarrow$.016} \\
Mongolia      & 0.882 & 0.848\,\textsubscript{$\downarrow$.034} \\
Slovakia      & 0.888 & 0.862\,\textsubscript{$\downarrow$.026} \\
Egypt         & 0.793 & 0.776\,\textsubscript{$\downarrow$.017} \\
\bottomrule
\end{tabular}
\caption{The alignment scores of the verbalized and balanced distributions for seven representative countries under no steering, averaged over the six models.}
\label{tab:supp_score_changes}
\end{table}

The two rankings agree on the broad ordering (Spearman $\rho$ between $0.88$ and $0.99$), so anchoring the distribution to the model's behavior does not overturn the country ranking; it corrects it locally.
The corrections are systematic.
Averaged over the six models, the country that falls the most is Mongolia ($+11.3$ ranks), followed by Mexico ($+9.0$), Thailand ($+7.3$), Serbia ($+6.7$), and Slovakia ($+6.3$); the largest single fall is Mongolia under Llama-3.1-70B-Instruct, from rank 15 to rank 39.
The countries that rise the most are the Netherlands ($-15.2$), Germany ($-12.2$), New Zealand ($-9.7$), the Philippines ($-8.0$), and Malaysia ($-6.5$).
Mongolia falls under all six models, from ranks 15, 18, 2, 15, 26, and 8 to ranks 39, 30, 10, 27, 32, and 14 for Llama-3.1-70B-Instruct, Qwen2.5-72B-Instruct, OLMo-3.1-32B-Instruct, GPT-4o, GPT-4.1, and DeepSeek-V4-Flash, respectively.
Under OLMo-3.1-32B-Instruct, in particular, the verbalized distribution places Mongolia second among all 66 countries, ahead of every English-speaking country except the United States, a position it loses once behavior is taken into account.

The countries at the two extremes of the ranking are essentially unaffected.
The United States and Northern Ireland remain in the top two under every model, and Myanmar, Egypt, Libya, and Iraq stay at the bottom: where the verbalized alignment score is genuinely high or genuinely low, there is no flatness artifact to remove.
The movement is concentrated in the middle of the ranking, exactly where a nearly uniform verbalized distribution can be rewarded for matching a flat human opinion distribution without predicting what the model would actually answer.

\begin{table*}[p]
\centering\scriptsize
\setlength{\tabcolsep}{1.5pt}
\begin{tabular}{l cc cc cc cc cc cc}
\toprule
\multirow{2}{*}{\textbf{Country}} & \multicolumn{2}{c}{\textbf{Llama}} & \multicolumn{2}{c}{\textbf{Qwen}} & \multicolumn{2}{c}{\textbf{OLMo}} & \multicolumn{2}{c}{\textbf{GPT-4o}} & \multicolumn{2}{c}{\textbf{GPT-4.1}} & \multicolumn{2}{c}{\textbf{DeepSeek}} \\
\cmidrule(lr){2-3}\cmidrule(lr){4-5}\cmidrule(lr){6-7}\cmidrule(lr){8-9}\cmidrule(lr){10-11}\cmidrule(lr){12-13}
& Verb & Ours & Verb & Ours & Verb & Ours & Verb & Ours & Verb & Ours & Verb & Ours \\
\midrule
United States & 0.906\textsubscript{1} & 0.889\textsubscript{3} & 0.905\textsubscript{1} & 0.902\textsubscript{1} & 0.892\textsubscript{1} & 0.875\textsubscript{3} & 0.917\textsubscript{1} & 0.903\textsubscript{1} & 0.927\textsubscript{1} & 0.918\textsubscript{1} & 0.911\textsubscript{1} & 0.903\textsubscript{1} \\
Northern Ireland & 0.902\textsubscript{2} & 0.899\textsubscript{1} & 0.902\textsubscript{2} & 0.896\textsubscript{2} & 0.876\textsubscript{8} & 0.863\textsubscript{5} & 0.908\textsubscript{2} & 0.893\textsubscript{2} & 0.921\textsubscript{2} & 0.912\textsubscript{2} & 0.899\textsubscript{5} & 0.889\textsubscript{2} \\
Chile & 0.890\textsubscript{4} & 0.856\textsubscript{15} & 0.899\textsubscript{4} & 0.870\textsubscript{11} & 0.887\textsubscript{3} & 0.861\textsubscript{7} & 0.908\textsubscript{3} & 0.881\textsubscript{7} & 0.896\textsubscript{7} & 0.878\textsubscript{11} & 0.908\textsubscript{2} & 0.886\textsubscript{4} \\
Canada & 0.884\textsubscript{7} & 0.890\textsubscript{2} & 0.899\textsubscript{3} & 0.892\textsubscript{3} & 0.884\textsubscript{5} & 0.880\textsubscript{1} & 0.902\textsubscript{4} & 0.892\textsubscript{3} & 0.906\textsubscript{3} & 0.897\textsubscript{4} & 0.892\textsubscript{9} & 0.888\textsubscript{3} \\
Slovakia & 0.878\textsubscript{11} & 0.851\textsubscript{19} & 0.884\textsubscript{12} & 0.853\textsubscript{20} & 0.881\textsubscript{7} & 0.852\textsubscript{15} & 0.896\textsubscript{5} & 0.875\textsubscript{9} & 0.887\textsubscript{12} & 0.862\textsubscript{20} & 0.900\textsubscript{4} & 0.880\textsubscript{6} \\
Argentina & 0.883\textsubscript{8} & 0.857\textsubscript{14} & 0.877\textsubscript{19} & 0.867\textsubscript{14} & 0.875\textsubscript{11} & 0.855\textsubscript{13} & 0.892\textsubscript{6} & 0.866\textsubscript{16} & 0.902\textsubscript{5} & 0.887\textsubscript{8} & 0.896\textsubscript{7} & 0.874\textsubscript{12} \\
Cyprus & 0.893\textsubscript{3} & 0.861\textsubscript{12} & 0.885\textsubscript{11} & 0.864\textsubscript{16} & 0.865\textsubscript{20} & 0.827\textsubscript{30} & 0.892\textsubscript{8} & 0.873\textsubscript{12} & 0.890\textsubscript{11} & 0.876\textsubscript{12} & 0.899\textsubscript{6} & 0.879\textsubscript{8} \\
Australia & 0.885\textsubscript{5} & 0.888\textsubscript{4} & 0.889\textsubscript{7} & 0.892\textsubscript{4} & 0.869\textsubscript{17} & 0.859\textsubscript{11} & 0.891\textsubscript{9} & 0.886\textsubscript{4} & 0.904\textsubscript{4} & 0.899\textsubscript{3} & 0.882\textsubscript{21} & 0.879\textsubscript{7} \\
Singapore & 0.885\textsubscript{6} & 0.865\textsubscript{9} & 0.893\textsubscript{5} & 0.881\textsubscript{8} & 0.869\textsubscript{15} & 0.858\textsubscript{12} & 0.886\textsubscript{12} & 0.876\textsubscript{8} & 0.881\textsubscript{18} & 0.868\textsubscript{19} & 0.889\textsubscript{10} & 0.877\textsubscript{9} \\
Hong Kong & 0.878\textsubscript{10} & 0.851\textsubscript{18} & 0.890\textsubscript{6} & 0.868\textsubscript{13} & 0.885\textsubscript{4} & 0.876\textsubscript{2} & 0.883\textsubscript{16} & 0.872\textsubscript{13} & 0.878\textsubscript{20} & 0.860\textsubscript{23} & 0.885\textsubscript{12} & 0.872\textsubscript{13} \\
Great Britain & 0.883\textsubscript{9} & 0.887\textsubscript{5} & 0.888\textsubscript{8} & 0.886\textsubscript{6} & 0.866\textsubscript{18} & 0.861\textsubscript{8} & 0.891\textsubscript{10} & 0.882\textsubscript{6} & 0.901\textsubscript{6} & 0.896\textsubscript{5} & 0.878\textsubscript{25} & 0.875\textsubscript{10} \\
Ukraine & 0.871\textsubscript{19} & 0.842\textsubscript{23} & 0.873\textsubscript{24} & 0.843\textsubscript{27} & 0.876\textsubscript{10} & 0.844\textsubscript{19} & 0.888\textsubscript{11} & 0.860\textsubscript{21} & 0.881\textsubscript{15} & 0.861\textsubscript{22} & 0.901\textsubscript{3} & 0.882\textsubscript{5} \\
Uruguay & 0.868\textsubscript{20} & 0.864\textsubscript{10} & 0.884\textsubscript{13} & 0.869\textsubscript{12} & 0.872\textsubscript{13} & 0.841\textsubscript{21} & 0.892\textsubscript{7} & 0.873\textsubscript{11} & 0.895\textsubscript{8} & 0.878\textsubscript{10} & 0.881\textsubscript{22} & 0.870\textsubscript{17} \\
Mongolia & 0.875\textsubscript{15} & 0.821\textsubscript{39} & 0.879\textsubscript{18} & 0.838\textsubscript{30} & 0.888\textsubscript{2} & 0.860\textsubscript{10} & 0.884\textsubscript{15} & 0.853\textsubscript{27} & 0.871\textsubscript{26} & 0.845\textsubscript{32} & 0.893\textsubscript{8} & 0.872\textsubscript{14} \\
Czech Republic & 0.867\textsubscript{22} & 0.844\textsubscript{21} & 0.881\textsubscript{16} & 0.849\textsubscript{23} & 0.883\textsubscript{6} & 0.862\textsubscript{6} & 0.885\textsubscript{13} & 0.870\textsubscript{14} & 0.874\textsubscript{23} & 0.849\textsubscript{29} & 0.883\textsubscript{15} & 0.868\textsubscript{20} \\
Taiwan & 0.872\textsubscript{16} & 0.864\textsubscript{11} & 0.884\textsubscript{14} & 0.872\textsubscript{10} & 0.860\textsubscript{24} & 0.848\textsubscript{16} & 0.885\textsubscript{14} & 0.882\textsubscript{5} & 0.880\textsubscript{19} & 0.868\textsubscript{18} & 0.883\textsubscript{14} & 0.875\textsubscript{11} \\
Andorra & 0.875\textsubscript{13} & 0.878\textsubscript{7} & 0.887\textsubscript{9} & 0.884\textsubscript{7} & 0.858\textsubscript{25} & 0.846\textsubscript{17} & 0.882\textsubscript{17} & 0.868\textsubscript{15} & 0.894\textsubscript{9} & 0.888\textsubscript{7} & 0.875\textsubscript{30} & 0.868\textsubscript{21} \\
New Zealand & 0.877\textsubscript{12} & 0.884\textsubscript{6} & 0.883\textsubscript{15} & 0.889\textsubscript{5} & 0.865\textsubscript{19} & 0.863\textsubscript{4} & 0.881\textsubscript{18} & 0.874\textsubscript{10} & 0.894\textsubscript{10} & 0.891\textsubscript{6} & 0.871\textsubscript{31} & 0.871\textsubscript{16} \\
Russia & 0.866\textsubscript{23} & 0.834\textsubscript{27} & 0.867\textsubscript{26} & 0.835\textsubscript{31} & 0.876\textsubscript{9} & 0.841\textsubscript{20} & 0.880\textsubscript{19} & 0.861\textsubscript{19} & 0.873\textsubscript{24} & 0.850\textsubscript{28} & 0.888\textsubscript{11} & 0.868\textsubscript{22} \\
Thailand & 0.868\textsubscript{21} & 0.838\textsubscript{25} & 0.885\textsubscript{10} & 0.846\textsubscript{24} & 0.870\textsubscript{14} & 0.830\textsubscript{28} & 0.878\textsubscript{21} & 0.854\textsubscript{25} & 0.861\textsubscript{31} & 0.838\textsubscript{37} & 0.883\textsubscript{16} & 0.869\textsubscript{18} \\
Brazil & 0.872\textsubscript{18} & 0.842\textsubscript{22} & 0.874\textsubscript{23} & 0.856\textsubscript{19} & 0.863\textsubscript{21} & 0.831\textsubscript{25} & 0.879\textsubscript{20} & 0.858\textsubscript{23} & 0.885\textsubscript{13} & 0.869\textsubscript{16} & 0.882\textsubscript{20} & 0.859\textsubscript{31} \\
Greece & 0.875\textsubscript{14} & 0.853\textsubscript{17} & 0.863\textsubscript{28} & 0.852\textsubscript{21} & 0.854\textsubscript{28} & 0.831\textsubscript{26} & 0.877\textsubscript{22} & 0.865\textsubscript{18} & 0.881\textsubscript{17} & 0.869\textsubscript{15} & 0.879\textsubscript{24} & 0.869\textsubscript{19} \\
Venezuela & 0.865\textsubscript{25} & 0.824\textsubscript{35} & 0.857\textsubscript{30} & 0.840\textsubscript{28} & 0.862\textsubscript{23} & 0.828\textsubscript{29} & 0.875\textsubscript{24} & 0.849\textsubscript{32} & 0.874\textsubscript{22} & 0.861\textsubscript{21} & 0.884\textsubscript{13} & 0.861\textsubscript{28} \\
Kazakhstan & 0.859\textsubscript{30} & 0.836\textsubscript{26} & 0.875\textsubscript{20} & 0.843\textsubscript{26} & 0.869\textsubscript{16} & 0.836\textsubscript{24} & 0.873\textsubscript{26} & 0.860\textsubscript{20} & 0.857\textsubscript{35} & 0.836\textsubscript{39} & 0.882\textsubscript{17} & 0.871\textsubscript{15} \\
Puerto Rico & 0.866\textsubscript{24} & 0.853\textsubscript{16} & 0.866\textsubscript{27} & 0.860\textsubscript{18} & 0.844\textsubscript{41} & 0.806\textsubscript{42} & 0.876\textsubscript{23} & 0.854\textsubscript{26} & 0.885\textsubscript{14} & 0.873\textsubscript{13} & 0.875\textsubscript{28} & 0.858\textsubscript{32} \\
Romania & 0.872\textsubscript{17} & 0.841\textsubscript{24} & 0.853\textsubscript{34} & 0.844\textsubscript{25} & 0.845\textsubscript{38} & 0.808\textsubscript{38} & 0.870\textsubscript{28} & 0.853\textsubscript{28} & 0.874\textsubscript{21} & 0.868\textsubscript{17} & 0.882\textsubscript{19} & 0.862\textsubscript{27} \\
Mexico & 0.862\textsubscript{27} & 0.815\textsubscript{48} & 0.852\textsubscript{35} & 0.829\textsubscript{35} & 0.855\textsubscript{27} & 0.817\textsubscript{36} & 0.874\textsubscript{25} & 0.841\textsubscript{39} & 0.869\textsubscript{27} & 0.853\textsubscript{25} & 0.880\textsubscript{23} & 0.853\textsubscript{35} \\
Guatemala & 0.861\textsubscript{28} & 0.816\textsubscript{47} & 0.849\textsubscript{38} & 0.830\textsubscript{34} & 0.857\textsubscript{26} & 0.820\textsubscript{33} & 0.870\textsubscript{30} & 0.844\textsubscript{36} & 0.868\textsubscript{28} & 0.854\textsubscript{24} & 0.882\textsubscript{18} & 0.862\textsubscript{26} \\
Germany & 0.863\textsubscript{26} & 0.871\textsubscript{8} & 0.874\textsubscript{21} & 0.877\textsubscript{9} & 0.849\textsubscript{34} & 0.840\textsubscript{23} & 0.869\textsubscript{31} & 0.866\textsubscript{17} & 0.881\textsubscript{16} & 0.878\textsubscript{9} & 0.860\textsubscript{41} & 0.860\textsubscript{30} \\
Macao SAR & 0.853\textsubscript{35} & 0.831\textsubscript{29} & 0.880\textsubscript{17} & 0.851\textsubscript{22} & 0.873\textsubscript{12} & 0.861\textsubscript{9} & 0.863\textsubscript{34} & 0.844\textsubscript{34} & 0.852\textsubscript{38} & 0.832\textsubscript{40} & 0.869\textsubscript{34} & 0.866\textsubscript{23} \\
Ecuador & 0.857\textsubscript{31} & 0.824\textsubscript{36} & 0.855\textsubscript{32} & 0.834\textsubscript{32} & 0.853\textsubscript{29} & 0.823\textsubscript{32} & 0.872\textsubscript{27} & 0.852\textsubscript{29} & 0.860\textsubscript{33} & 0.846\textsubscript{30} & 0.876\textsubscript{27} & 0.858\textsubscript{34} \\
Japan & 0.860\textsubscript{29} & 0.850\textsubscript{20} & 0.874\textsubscript{22} & 0.863\textsubscript{17} & 0.852\textsubscript{30} & 0.840\textsubscript{22} & 0.863\textsubscript{35} & 0.852\textsubscript{30} & 0.864\textsubscript{30} & 0.852\textsubscript{26} & 0.859\textsubscript{42} & 0.852\textsubscript{36} \\
Morocco & 0.856\textsubscript{32} & 0.828\textsubscript{30} & 0.850\textsubscript{37} & 0.826\textsubscript{40} & 0.849\textsubscript{35} & 0.806\textsubscript{40} & 0.865\textsubscript{32} & 0.843\textsubscript{38} & 0.859\textsubscript{34} & 0.844\textsubscript{33} & 0.877\textsubscript{26} & 0.860\textsubscript{29} \\
South Korea & 0.852\textsubscript{37} & 0.823\textsubscript{37} & 0.868\textsubscript{25} & 0.839\textsubscript{29} & 0.862\textsubscript{22} & 0.854\textsubscript{14} & 0.861\textsubscript{37} & 0.844\textsubscript{37} & 0.848\textsubscript{41} & 0.830\textsubscript{41} & 0.868\textsubscript{35} & 0.852\textsubscript{37} \\
Kenya & 0.850\textsubscript{39} & 0.827\textsubscript{31} & 0.857\textsubscript{31} & 0.830\textsubscript{33} & 0.850\textsubscript{32} & 0.819\textsubscript{34} & 0.870\textsubscript{29} & 0.852\textsubscript{31} & 0.852\textsubscript{39} & 0.837\textsubscript{38} & 0.875\textsubscript{29} & 0.863\textsubscript{25} \\
Serbia & 0.850\textsubscript{40} & 0.812\textsubscript{52} & 0.844\textsubscript{41} & 0.819\textsubscript{48} & 0.849\textsubscript{33} & 0.818\textsubscript{35} & 0.864\textsubscript{33} & 0.839\textsubscript{42} & 0.861\textsubscript{32} & 0.843\textsubscript{34} & 0.870\textsubscript{32} & 0.850\textsubscript{40} \\
Colombia & 0.853\textsubscript{36} & 0.815\textsubscript{49} & 0.845\textsubscript{39} & 0.827\textsubscript{38} & 0.846\textsubscript{37} & 0.807\textsubscript{39} & 0.862\textsubscript{36} & 0.841\textsubscript{40} & 0.865\textsubscript{29} & 0.850\textsubscript{27} & 0.868\textsubscript{37} & 0.843\textsubscript{46} \\
Netherlands & 0.855\textsubscript{33} & 0.858\textsubscript{13} & 0.860\textsubscript{29} & 0.866\textsubscript{15} & 0.844\textsubscript{40} & 0.845\textsubscript{18} & 0.855\textsubscript{41} & 0.856\textsubscript{24} & 0.871\textsubscript{25} & 0.869\textsubscript{14} & 0.848\textsubscript{48} & 0.848\textsubscript{41} \\
Malaysia & 0.846\textsubscript{43} & 0.827\textsubscript{32} & 0.852\textsubscript{36} & 0.827\textsubscript{37} & 0.850\textsubscript{31} & 0.831\textsubscript{27} & 0.860\textsubscript{38} & 0.858\textsubscript{22} & 0.837\textsubscript{45} & 0.822\textsubscript{45} & 0.870\textsubscript{33} & 0.864\textsubscript{24} \\
Bolivia & 0.854\textsubscript{34} & 0.818\textsubscript{44} & 0.841\textsubscript{44} & 0.828\textsubscript{36} & 0.843\textsubscript{42} & 0.812\textsubscript{37} & 0.858\textsubscript{39} & 0.844\textsubscript{35} & 0.856\textsubscript{36} & 0.845\textsubscript{31} & 0.868\textsubscript{36} & 0.848\textsubscript{42} \\
Nicaragua & 0.845\textsubscript{44} & 0.808\textsubscript{55} & 0.844\textsubscript{40} & 0.823\textsubscript{42} & 0.848\textsubscript{36} & 0.806\textsubscript{41} & 0.857\textsubscript{40} & 0.830\textsubscript{48} & 0.856\textsubscript{37} & 0.840\textsubscript{36} & 0.864\textsubscript{38} & 0.847\textsubscript{44} \\
Turkey & 0.851\textsubscript{38} & 0.823\textsubscript{38} & 0.855\textsubscript{33} & 0.827\textsubscript{39} & 0.840\textsubscript{43} & 0.797\textsubscript{48} & 0.850\textsubscript{44} & 0.840\textsubscript{41} & 0.835\textsubscript{46} & 0.821\textsubscript{46} & 0.859\textsubscript{43} & 0.850\textsubscript{39} \\
Lebanon & 0.850\textsubscript{41} & 0.807\textsubscript{56} & 0.838\textsubscript{46} & 0.819\textsubscript{46} & 0.834\textsubscript{44} & 0.799\textsubscript{46} & 0.847\textsubscript{45} & 0.831\textsubscript{47} & 0.840\textsubscript{42} & 0.829\textsubscript{43} & 0.860\textsubscript{40} & 0.847\textsubscript{45} \\
Peru & 0.848\textsubscript{42} & 0.811\textsubscript{53} & 0.828\textsubscript{51} & 0.819\textsubscript{45} & 0.826\textsubscript{48} & 0.800\textsubscript{45} & 0.847\textsubscript{46} & 0.838\textsubscript{43} & 0.849\textsubscript{40} & 0.842\textsubscript{35} & 0.862\textsubscript{39} & 0.840\textsubscript{48} \\
Philippines & 0.827\textsubscript{52} & 0.826\textsubscript{34} & 0.843\textsubscript{43} & 0.825\textsubscript{41} & 0.845\textsubscript{39} & 0.825\textsubscript{31} & 0.852\textsubscript{42} & 0.848\textsubscript{33} & 0.825\textsubscript{49} & 0.812\textsubscript{50} & 0.854\textsubscript{45} & 0.858\textsubscript{33} \\
India & 0.830\textsubscript{51} & 0.818\textsubscript{42} & 0.843\textsubscript{42} & 0.809\textsubscript{50} & 0.832\textsubscript{46} & 0.786\textsubscript{51} & 0.851\textsubscript{43} & 0.835\textsubscript{45} & 0.827\textsubscript{48} & 0.808\textsubscript{54} & 0.855\textsubscript{44} & 0.850\textsubscript{38} \\
Zimbabwe & 0.838\textsubscript{46} & 0.827\textsubscript{33} & 0.837\textsubscript{47} & 0.819\textsubscript{47} & 0.820\textsubscript{50} & 0.787\textsubscript{50} & 0.846\textsubscript{47} & 0.833\textsubscript{46} & 0.838\textsubscript{43} & 0.830\textsubscript{42} & 0.849\textsubscript{47} & 0.841\textsubscript{47} \\
Iran & 0.840\textsubscript{45} & 0.832\textsubscript{28} & 0.836\textsubscript{48} & 0.817\textsubscript{49} & 0.830\textsubscript{47} & 0.791\textsubscript{49} & 0.845\textsubscript{48} & 0.836\textsubscript{44} & 0.831\textsubscript{47} & 0.819\textsubscript{47} & 0.852\textsubscript{46} & 0.848\textsubscript{43} \\
Nigeria & 0.832\textsubscript{49} & 0.817\textsubscript{46} & 0.833\textsubscript{50} & 0.808\textsubscript{51} & 0.820\textsubscript{51} & 0.780\textsubscript{54} & 0.840\textsubscript{49} & 0.823\textsubscript{50} & 0.825\textsubscript{50} & 0.813\textsubscript{49} & 0.846\textsubscript{49} & 0.838\textsubscript{49} \\
Armenia & 0.837\textsubscript{47} & 0.809\textsubscript{54} & 0.826\textsubscript{52} & 0.805\textsubscript{53} & 0.816\textsubscript{54} & 0.773\textsubscript{56} & 0.833\textsubscript{52} & 0.815\textsubscript{57} & 0.837\textsubscript{44} & 0.827\textsubscript{44} & 0.842\textsubscript{50} & 0.825\textsubscript{58} \\
Uzbekistan & 0.819\textsubscript{55} & 0.818\textsubscript{41} & 0.840\textsubscript{45} & 0.821\textsubscript{43} & 0.832\textsubscript{45} & 0.803\textsubscript{44} & 0.836\textsubscript{50} & 0.825\textsubscript{49} & 0.820\textsubscript{54} & 0.811\textsubscript{52} & 0.832\textsubscript{57} & 0.833\textsubscript{53} \\
China & 0.822\textsubscript{54} & 0.819\textsubscript{40} & 0.836\textsubscript{49} & 0.820\textsubscript{44} & 0.822\textsubscript{49} & 0.798\textsubscript{47} & 0.826\textsubscript{57} & 0.811\textsubscript{58} & 0.820\textsubscript{53} & 0.807\textsubscript{55} & 0.831\textsubscript{58} & 0.829\textsubscript{55} \\
Maldives & 0.830\textsubscript{50} & 0.817\textsubscript{45} & 0.819\textsubscript{57} & 0.801\textsubscript{55} & 0.805\textsubscript{61} & 0.767\textsubscript{59} & 0.833\textsubscript{51} & 0.823\textsubscript{51} & 0.820\textsubscript{52} & 0.811\textsubscript{51} & 0.840\textsubscript{51} & 0.834\textsubscript{51} \\
Tunisia & 0.835\textsubscript{48} & 0.795\textsubscript{59} & 0.814\textsubscript{59} & 0.798\textsubscript{58} & 0.808\textsubscript{58} & 0.767\textsubscript{58} & 0.825\textsubscript{58} & 0.805\textsubscript{59} & 0.823\textsubscript{51} & 0.813\textsubscript{48} & 0.835\textsubscript{53} & 0.819\textsubscript{61} \\
Indonesia & 0.819\textsubscript{56} & 0.818\textsubscript{43} & 0.823\textsubscript{55} & 0.807\textsubscript{52} & 0.815\textsubscript{55} & 0.783\textsubscript{53} & 0.830\textsubscript{53} & 0.815\textsubscript{56} & 0.813\textsubscript{57} & 0.806\textsubscript{57} & 0.834\textsubscript{54} & 0.829\textsubscript{54} \\
Tajikistan & 0.819\textsubscript{58} & 0.814\textsubscript{51} & 0.825\textsubscript{53} & 0.801\textsubscript{56} & 0.818\textsubscript{53} & 0.785\textsubscript{52} & 0.828\textsubscript{55} & 0.819\textsubscript{52} & 0.807\textsubscript{59} & 0.796\textsubscript{59} & 0.837\textsubscript{52} & 0.836\textsubscript{50} \\
Ethiopia & 0.825\textsubscript{53} & 0.815\textsubscript{50} & 0.824\textsubscript{54} & 0.802\textsubscript{54} & 0.806\textsubscript{59} & 0.763\textsubscript{60} & 0.828\textsubscript{56} & 0.818\textsubscript{53} & 0.817\textsubscript{55} & 0.808\textsubscript{53} & 0.834\textsubscript{55} & 0.825\textsubscript{57} \\
Kyrgyzstan & 0.819\textsubscript{57} & 0.805\textsubscript{57} & 0.819\textsubscript{56} & 0.799\textsubscript{57} & 0.809\textsubscript{57} & 0.769\textsubscript{57} & 0.829\textsubscript{54} & 0.816\textsubscript{55} & 0.816\textsubscript{56} & 0.806\textsubscript{56} & 0.834\textsubscript{56} & 0.828\textsubscript{56} \\
Iraq & 0.813\textsubscript{61} & 0.778\textsubscript{65} & 0.805\textsubscript{61} & 0.779\textsubscript{61} & 0.811\textsubscript{56} & 0.774\textsubscript{55} & 0.822\textsubscript{59} & 0.795\textsubscript{62} & 0.811\textsubscript{58} & 0.792\textsubscript{60} & 0.829\textsubscript{59} & 0.819\textsubscript{60} \\
Viet Nam & 0.791\textsubscript{64} & 0.798\textsubscript{58} & 0.816\textsubscript{58} & 0.796\textsubscript{59} & 0.819\textsubscript{52} & 0.806\textsubscript{43} & 0.821\textsubscript{60} & 0.817\textsubscript{54} & 0.796\textsubscript{63} & 0.782\textsubscript{64} & 0.822\textsubscript{61} & 0.833\textsubscript{52} \\
Pakistan & 0.814\textsubscript{59} & 0.793\textsubscript{60} & 0.813\textsubscript{60} & 0.781\textsubscript{60} & 0.806\textsubscript{60} & 0.757\textsubscript{61} & 0.821\textsubscript{61} & 0.800\textsubscript{60} & 0.803\textsubscript{61} & 0.787\textsubscript{63} & 0.826\textsubscript{60} & 0.820\textsubscript{59} \\
Jordan & 0.814\textsubscript{60} & 0.787\textsubscript{61} & 0.792\textsubscript{62} & 0.776\textsubscript{62} & 0.788\textsubscript{62} & 0.746\textsubscript{63} & 0.806\textsubscript{62} & 0.795\textsubscript{61} & 0.799\textsubscript{62} & 0.787\textsubscript{62} & 0.813\textsubscript{62} & 0.807\textsubscript{63} \\
Libya & 0.808\textsubscript{63} & 0.784\textsubscript{63} & 0.788\textsubscript{65} & 0.776\textsubscript{63} & 0.787\textsubscript{63} & 0.746\textsubscript{64} & 0.803\textsubscript{63} & 0.790\textsubscript{64} & 0.804\textsubscript{60} & 0.798\textsubscript{58} & 0.810\textsubscript{63} & 0.799\textsubscript{64} \\
Egypt & 0.812\textsubscript{62} & 0.784\textsubscript{64} & 0.789\textsubscript{64} & 0.774\textsubscript{64} & 0.772\textsubscript{65} & 0.736\textsubscript{65} & 0.796\textsubscript{64} & 0.785\textsubscript{65} & 0.794\textsubscript{64} & 0.788\textsubscript{61} & 0.798\textsubscript{65} & 0.787\textsubscript{66} \\
Bangladesh & 0.788\textsubscript{65} & 0.786\textsubscript{62} & 0.791\textsubscript{63} & 0.770\textsubscript{65} & 0.779\textsubscript{64} & 0.748\textsubscript{62} & 0.792\textsubscript{65} & 0.791\textsubscript{63} & 0.776\textsubscript{65} & 0.768\textsubscript{65} & 0.801\textsubscript{64} & 0.807\textsubscript{62} \\
Myanmar & 0.772\textsubscript{66} & 0.770\textsubscript{66} & 0.783\textsubscript{66} & 0.760\textsubscript{66} & 0.767\textsubscript{66} & 0.729\textsubscript{66} & 0.784\textsubscript{66} & 0.778\textsubscript{66} & 0.766\textsubscript{66} & 0.753\textsubscript{66} & 0.792\textsubscript{66} & 0.790\textsubscript{65} \\
\bottomrule
\end{tabular}
\caption{Alignment score and country ranking of the verbalized distribution (Verb) and the balanced distribution (Ours) on the 66 WVS countries and regions, under no steering. The subscript is the rank of the country among the 66 under that distribution (1 = highest alignment score). The balanced distribution uses the persona-steering weights of Table~\ref{tab:persona_weights}, averaged over the three datasets, applied unchanged to the WVS distributions. Countries are sorted by their average verbalized rank over the six models.}
\label{tab:supp_mongolia_rank}
\end{table*}

\section{External Validation with a Cultural Knowledge Benchmark}
\label{sec:supp_blend}
Section~\ref{sec:mongolia} argues that anchoring the distribution to the model's behavior corrects the country ranking rather than merely changing it.
This section describes the external test of that claim and reports the full results.

\paragraph{Benchmark.}
BLEnD~\citep{myung2024blend} measures everyday cultural knowledge (food, celebrations, education, work, and daily life) with questions written and answered by native annotators.
Eleven of its countries and regions map onto the 66 WVS groups: the United States, Great Britain, China, Greece, Mexico, Iran, South Korea, Ethiopia, Indonesia (West Java), India (Assam), and Nigeria (Northern Nigeria).

\paragraph{Tasks and Scoring.}
We evaluate the six models of Section~\ref{sec:setup} on the two English tasks.
For multiple-choice QA, we sample 2,300 questions per country with a fixed seed, matching the per country scale of the original benchmark, and score the emitted option letter against the gold answer.
For short-answer QA, we use all 500 questions per country and follow BLEnD's answer-containment rule: a response counts as correct if it contains any annotator answer.
The local-language short-answer split is excluded because BLEnD provides no local-language multiple-choice task, the United States and Great Britain have no local-language split, and only nine scorable countries would remain.
Open-weight models are run with greedy decoding; GPT-4o and GPT-4.1 are queried through the Batch API.

\paragraph{Correlation.}
For every model and task, we compute the Pearson correlation between the 11 countries' BLEnD accuracy and their WVS alignment scores under the log-probability, verbalized, and balanced distributions.
The setting is that of \S\ref{sec:supp_mongolia}: no steering, with the balanced distribution using the persona-steering weights of Table~\ref{tab:persona_weights} averaged over the three datasets.
Table~\ref{tab:blend} reports all 12 model and task pairs.
The balanced distribution correlates better than the verbalized one in 10 of the 12 pairs (two-sided sign test, $p=0.039$).
The conclusion does not depend on the choice of statistic: Kendall's $\tau$ also prefers the balanced distribution in 10 of 12 pairs ($p=0.039$), and Spearman's $\rho$ in 9 of 12.
The two exceptions are GPT-4o and GPT-4.1 on multiple-choice QA, the two models with the smallest transferred log-probability weight $w_{lp}$ ($0.41$ and $0.23$): their balanced distribution stays closest to their verbalized one and has the least room to move.
The log-probability distribution attains the highest correlation in four of the pairs.
As it is the model's behavior itself, this corroborates that the gain of the balanced distribution comes from anchoring to the behavioral signal.
Unlike the raw log-probability distribution, the balanced distribution preserves the distribution quality reported in the main results.
\end{document}